%% file: group_sparse_neurips.tex
\documentclass{article}
\pdfoutput=1

    \usepackage[final, nonatbib]{neurips_2020}

\usepackage[utf8]{inputenc} 
\usepackage[T1]{fontenc}    
\usepackage{hyperref}       
\usepackage{url}            
\usepackage{booktabs}       
\usepackage{amsfonts}       
\usepackage{nicefrac}       
\usepackage{microtype}      

\usepackage{epsfig}
\usepackage{graphicx}
\usepackage{amsmath}
\usepackage{amssymb}

\usepackage[dvipsnames]{xcolor}
\usepackage{subfigure}
\usepackage{helvet}
\usepackage{courier}
\usepackage{bm}
\usepackage{dsfont}
\usepackage{paralist}

\usepackage{multirow}
\usepackage{wrapfig}
\usepackage{mathtools,cuted}

\input{taesup.sty}

\newcommand{\eg}{{\it e.g.}}
\newcommand{\ie}{{\it i.e.}}

\newcommand{\btheta}{\bm \theta}
\newcommand{\bthetanl}{\bm\theta_{n_\ell}}
\newcommand{\hatbtheta}{\hat{\bm\theta}_{n_\ell}^{(t-1)}}
\newcommand{\bprox}{\textbf{prox}}
\newcommand{\olm}{\Omega_{n_\ell}^{t-1}}
\newcommand{\Gcalz}{\mathcal{G}_0}

\title{Continual Learning with Node-Importance based Adaptive Group Sparse Regularization}

\author{%
  Sangwon Jung\textsuperscript{\rm 1}\thanks{Equal contribution.},\ \ Hongjoon Ahn\textsuperscript{\rm 2}\footnotemark[1],\ \ Sungmin Cha\textsuperscript{\rm 1}  and Taesup Moon\textsuperscript{\rm 1,2} \\
  \textsuperscript{\rm 1}Department of Electrical and Computer Engineering, 
  \textsuperscript{\rm 2} Department of Artificial Intelligence, \\
  Sungkyunkwan University, Suwon, Korea 16419\\
  \texttt{\{s.jung, hong0805, csm9493, tsmoon\}@skku.edu}
}

\begin{document}

\maketitle

\begin{abstract}
  We propose a novel regularization-based continual learning method, dubbed as Adaptive Group Sparsity based Continual Learning (AGS-CL), using two group sparsity-based penalties. Our method selectively employs the two penalties when learning each neural network \textit{node} based on its the importance, which is adaptively updated after learning each task. By utilizing the proximal gradient descent method, the exact sparsity and freezing of the model is guaranteed during the learning process, and thus, the learner explicitly controls the model capacity. Furthermore, as a critical detail, we re-initialize the weights associated with unimportant nodes after learning each task in order to facilitate efficient learning and prevent the negative transfer. Throughout the extensive experimental results, we show that our AGS-CL uses orders of magnitude less memory space for storing the regularization parameters, and it significantly outperforms several state-of-the-art baselines on representative benchmarks for both supervised and reinforcement learning. 


\end{abstract}

\input{intro}

\input{main}

\input{experiments}
\input{conclusion}
\bibliographystyle{plain}
\bibliography{bibfile}

\end{document}


\maketitle

\section{Proof of Lemma 1}
 From (Eq.(3), manuscript), $\bprox_{\alpha f}(\bm v)$ minimizes the convex function
\begin{align}
    \ell(\btheta)\triangleq c\|\btheta-\btheta_0\|_2+\frac{1}{2\alpha}\|\btheta-\bm v\|_2^2,\label{eq:prox_loss}
\end{align}
and for brevity, denote $\btheta^*:=\bprox_{\alpha f}(\bm v)$ as the minimizer. Denoting $\partial_{\btheta}\ell(\btheta)$ as the set of subgradients of $\ell(\btheta)$, we know that $\btheta^*\in\{\btheta:\partial_{\btheta}\ell(\btheta)=0\}$ since $\ell(\btheta)$ is convex. 
Also, by denoting $\bm w$ as the subgradient of $\|\btheta-\btheta_0\|_2$ at $\btheta^*$,
we then have the optimality condition,
\begin{align}
     \frac{1}{\alpha}(\bm v - \btheta^*)= c\bm w.\label{eq:subgradient0}
\end{align}
Since $\|\btheta-\btheta_0\|_2$ is not differentiable at $\btheta=\btheta_0$, we know  
\begin{equation}
\label{eq : subgradient w}
\bm w =
    \begin{cases*}
      \frac{\btheta^*-\btheta_0}{\norm{\btheta^*-\btheta_0}_2} & if $\btheta^* \neq \btheta_0$ \\
      \in \{\bm w:\norm{\bm w}_2<1\}        & if $\btheta^* = \btheta_0$
    \end{cases*}.
\end{equation}
Now, taking $\ell_2$-norm on both sides of (\ref{eq:subgradient0}), we can deduce
\begin{align}
\btheta^*=\btheta_0 \ \ \text{if and only if} \ \   \norm{\bm v - \btheta^*}_2 < \alpha c.\label{eq:sol_1}
\end{align}
Moreover, if $\btheta^*\neq\btheta_0$, we can derive from (\ref{eq:subgradient0}) and (\ref{eq : subgradient w}) that 
\begin{align}
    \|\bm v-\btheta_0\|_2-\alpha c=\|\btheta^*-\btheta\|_2\geq 0,
\end{align}
and correspondingly, 
\begin{align}
    \btheta^* = \Big( 1-\frac{\alpha c}{\|\btheta_0-\bm v\|_2}\Big)\bm v+\frac{\alpha c}{\|\btheta_0-\bm v\|_2}\btheta_0.\label{eq:sol2}
\end{align}
Combining (\ref{eq:sol_1}) and (\ref{eq:sol2}), we have the lemma. \ \ \qed

\section{Additional ablation studies}

\subsection{Ablation study of $\rho$}

\begin{figure}[h]
    \centering
    \includegraphics[width=0.4\textwidth]{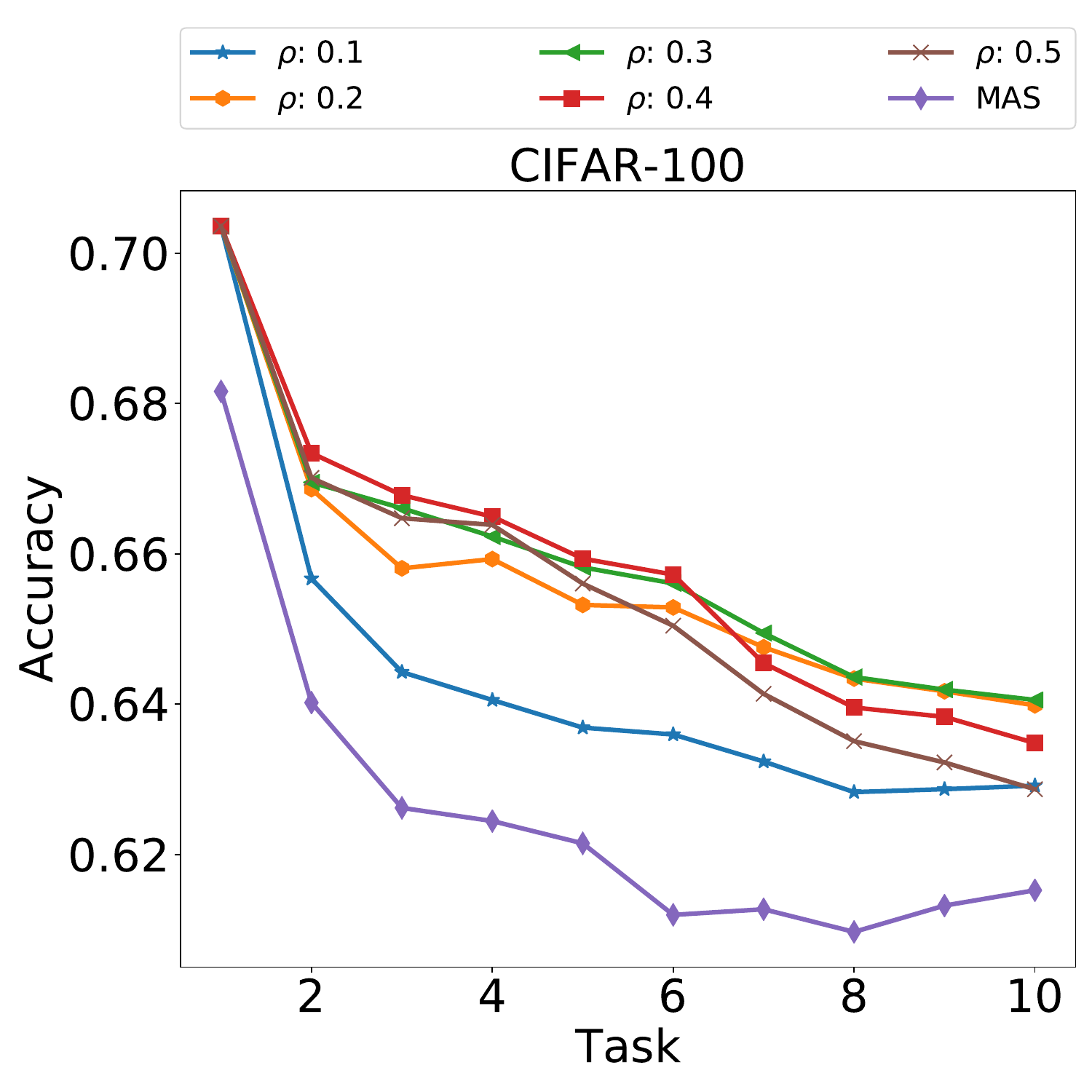}
    \vspace{-.1in}
    \caption{Average accuracy of AGS-CL on CIFAR-100 depending on $\rho$}\label{fig:gamma-cifar}
    
\end{figure}

Here, we analyze the effect of $\rho$ for the \textbf{[Rand-init]} described in Section 3.4 (manuscript) (I.2). Figure \ref{fig:gamma-cifar} below reports the average accuracy on CIFAR-100 for AGS-CL and MAS. For AGS-CL, we fixed $(\mu, \lambda)=(10,400)$ and varied $\rho\in\{0.1,\ldots,0.5\}$, and for MAS, we used the optimal hyperparameter. First, we observe that for $\rho\leq0.5$, AGS-CL is not very sensitive to $\rho$, and it outperforms MAS for all $\rho$. 
Second, we observe that $\rho$ affects the plasticity for learning new tasks. Namely, while $\rho=0.1$ and $\rho=0.5$ achieve the same final average accuracy, we note $\rho=0.1$ suffers earlier since it does not sufficiently grow the network capacity for learning new tasks, whereas $\rho=0.5$ suffers later since it uses up the network capacity too much in early tasks and makes the network too stable for later tasks. Thus, appropriate $\rho$ may find the right trade-off between the sparsity and the used capacity of the network and achieve higher average accuracy.


\subsection{Effect of PGD updates}

\begin{figure}[h]
    \centering
    \subfigure[Average accuracy with and without PGD.]{\label{fig:pgd-nopgd-accuracy}
    \includegraphics[width=0.48\textwidth]{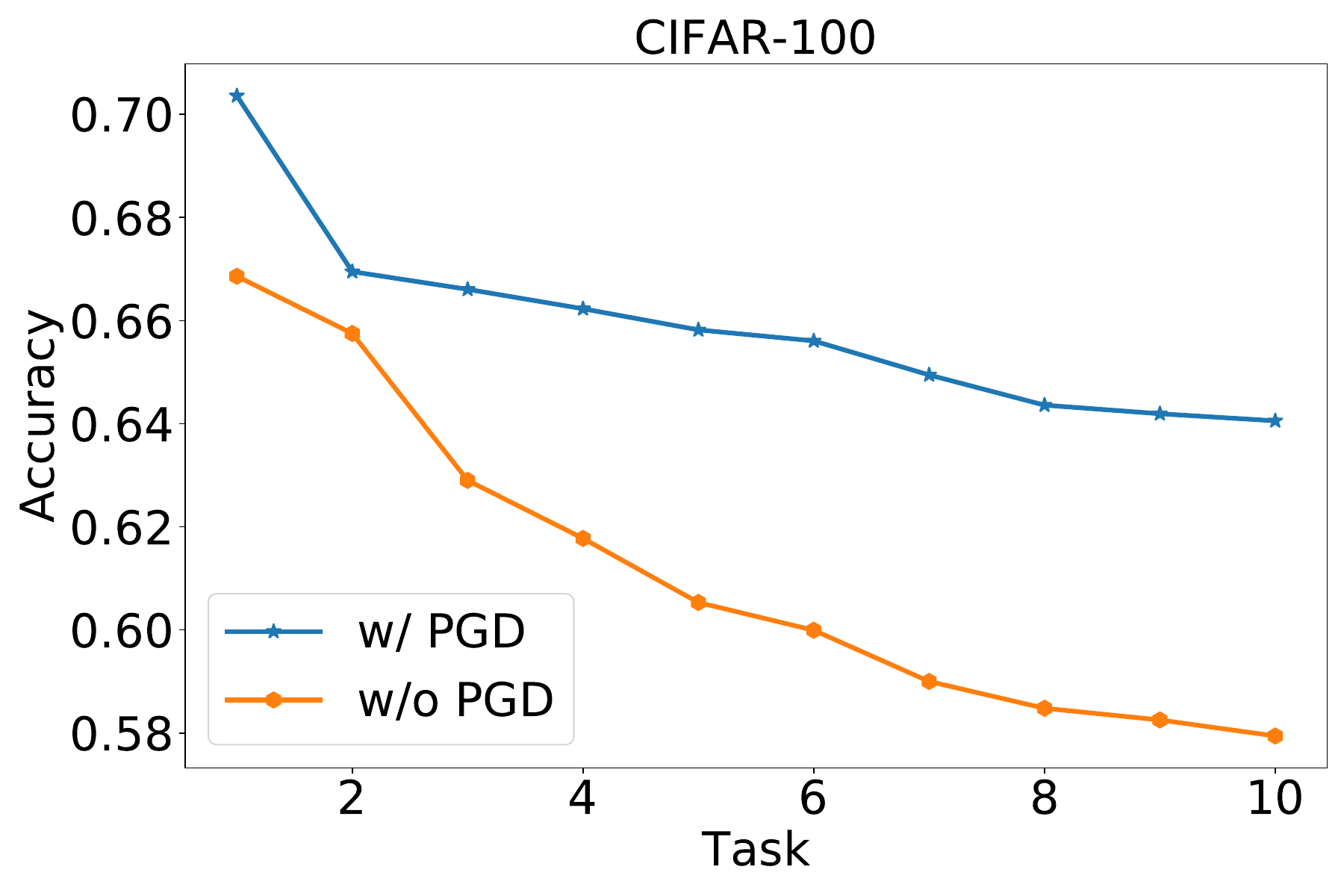}}
    \subfigure[Sparsity (decreasing curves) and used capacity (increasing curves) with and without PGD.]{\label{fig:pgd-nopgd-sparsity}
    \includegraphics[width=0.48\textwidth]{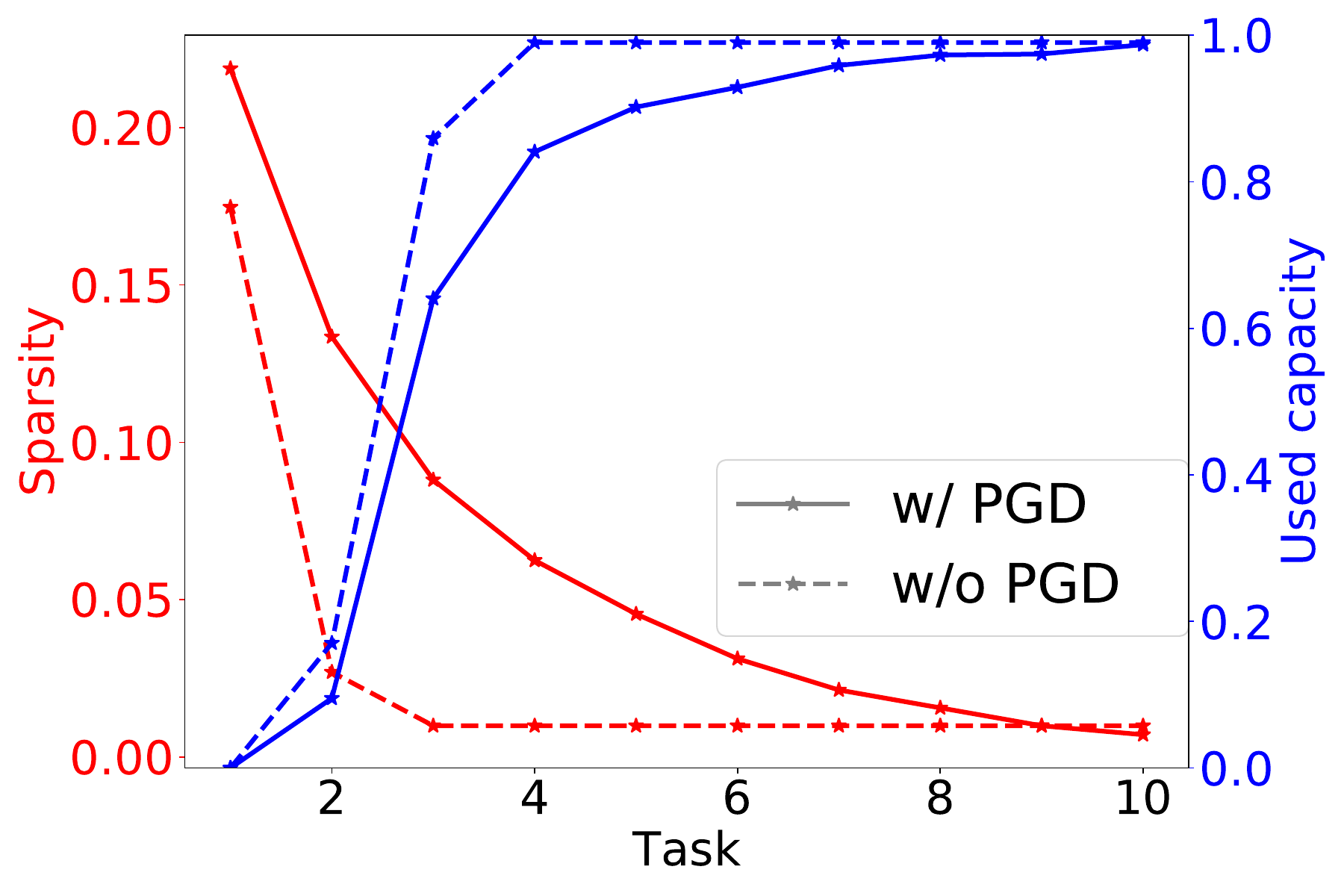}}
    \vspace{-.1in}
    \caption{Ablation study on PGD for CIFAR-100}\label{fig:pgd-nopgd}
    
\end{figure}

As mentioned in Section 3.3 (manuscript), our PGD update plays a critical role in achieving high accuracy. Here, we compare with a method without PGD. 
Figure \ref{fig:pgd-nopgd-accuracy} and Figure \ref{fig:pgd-nopgd-sparsity} show the average accuracy and the sparsity and used capacity on CIFAR-100. `\textit{w/o} PGD' in Figure \ref{fig:pgd-nopgd} indicates training the network without PGD, \textit{i.e.}, the Adam step was used for optimizing $ \mathcal{L}_t(\bm\theta)$ (Eq.(2), manuscript)  which implies the combined loss of $\mathcal{L}_{\text{TS},t}(\btheta)$ and group sparse regularizations(term (a) and term (b) of Eq.(2), manuscript). Since optimizing $\mathcal{L}_t(\bm\theta)$ using Adam cannot achieve the global optimal point of group sparse regularization, we used a proper threshold $\tau$ to modify the definition of $\Gcalz$ in (Eq.(1), manuscript) and the used capacity.
Thus, we define $\Gcalz^{t-1}\triangleq\{n_\ell:\olm<\tau\}\subseteq\Gcal$, and used capacity as $|\{n_\ell: \|\hat{\btheta}_{n_\ell}^{(t)}-\hat{\btheta}_{n_\ell}^{(t-1)}\|_2<\tau\}|/|\Gcal|$. Except for above definitions, all the common hyperparameters and training settings are same as `\textit{w/} PGD', and we set the threshold $\tau=10^{-4}$.

Followings are our observations. First, the average accuracy (Figure \ref{fig:pgd-nopgd-accuracy}) of `\textit{w/o} PGD' is much lower than `\textit{w/} PGD', which indicates that our PGD updates not only require \textit{less} hyperparameters (\textit{i.e.}, does not need $\tau$ threshold), but also does a much more accurate sparsification and freezing for achieving high accuracy. 
Second, we observe the sparsity (Figure \ref{fig:pgd-nopgd-sparsity}) of `\textit{w/o} PGD' decreases much faster than `\textit{w/} PGD'. The reason is because the weights associated with the nodes in $\Gcalz^{t}$ are not exactly zero, hence, the gradients for those weights do not vanish, which cause the unimportant nodes in $\Gcalz^{t}$ also continuously learn in every task. From these results, we conclude our PGD update is essential in AGS-CL. 
\subsection{Comparison with EWC}

\begin{wrapfigure}{r}{0.38\textwidth}
    \centering
    \vspace{-.3in}
    \includegraphics[width=0.38\textwidth]{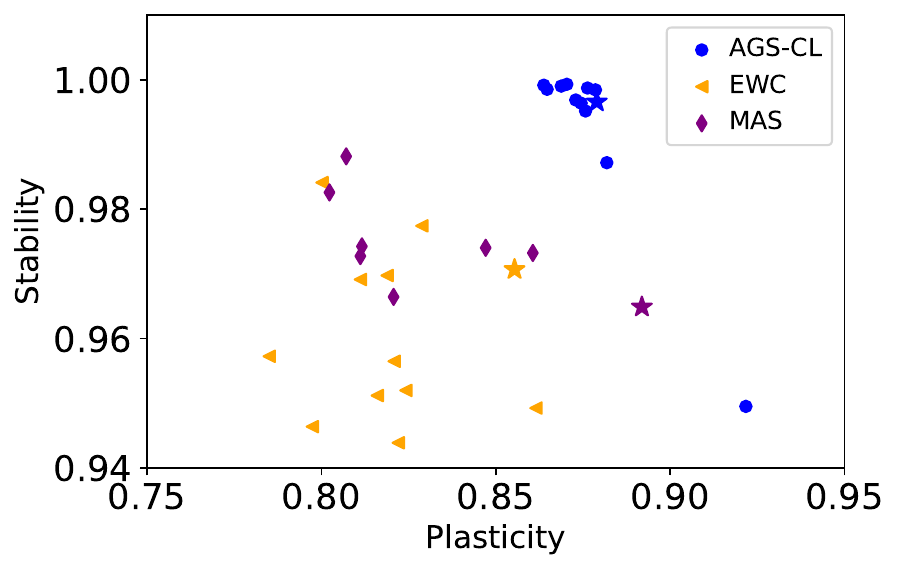}
    \vspace{-.23in}
    \caption{Plasticity ($\mathcal{P}$) and stability ($\mathcal{S}$) for CIFAR-100}
    \label{fig:init_ablation_ewc}
    \vspace{-.2in}
\end{wrapfigure}
We additionally evaluate the performance of EWC with two measures, plasticity ($\mathcal{P}$) and stability ($\mathcal{S}$), which are proposed in (Figure 5(c), manuscript). 
Figure \ref{fig:init_ablation_ewc} reports the trade-offs between $\mathcal{P}$ and $\mathcal{S}$ for AGS-CL, MAS and EWC. The plotted trade-offs of EWC are over the $\lambda$ and the others are the same as (Figure 5(c), manuscript). Note that although EWC has comparable $\mathcal{P}$-$\mathcal{S}$ trade-offs with MAS, AGS-CL apparently has the better $\mathcal{P}$-$\mathcal{S}$ trade-offs than EWC and MAS.

\section{Implementation details}

\subsection{Supervised learning}
In CIFAR-100, CIFAR-10/100 and Omniglot \textcolor{magenta}{\footnote{\textcolor{magenta}{https://drive.google.com/file/d/1WxFZQyt3v7QRHwxFbdb1KO02XWLT0R9z/view?usp=sharing}}}, we train all methods with mini-batch size of 256 for 100 epochs using Adam optimizer \cite{KinBa15} with initial learning rate 0.001 and decaying it by a factor of 3 if there is no improvement in the validation loss for 5 consecutive epochs, similarly as in \cite{SerraSurisMironKarat2018(HAT)}. In CUB200\textcolor{magenta}{\footnote{\textcolor{magenta}{{https://github.com/visipedia/tf\_classification/wiki/CUB-200-Image-Classification}}}}, we train all methods with mini-batch size 64 for 40 epochs using SGD with momentum 0.9 with initial learning rate 0.005 and decay it by a factor of 10 after training 30 epochs.

\subsubsection{Hyperparameters for supervised learning experiments}
The details on hyperparameters are in Table \ref{table:hyperparameter_supervised}. For AGS-CL, we set $\eta$ to 0.9 and for RWALK, we set $\alpha$ to 0.9 for all datasets. We extensively searched the best hyperparameter for each method to make the comparison as fair as possible. 

\begin{table}[ht]
\small
\centering
\caption{Hyperparameters for supervised learning experiments}
\resizebox{1.0\textwidth}{!}{\begin{tabular}{c|c|c|c|c|c}
\hline
Methods\textbackslash{}Dataset & CIFAR-100                                                                        & CIFAR-10/100                                                                      & Omniglot                                                                          & CUB200                                                                            & \begin{tabular}[c]{@{}c@{}}Sequence of \\ 8 different datasets\end{tabular}                                                       \\ \hline \hline
AGS-CL                         & \begin{tabular}[c]{@{}c@{}}$\lambda$ (400)\\ $\mu$(10), $\rho$(0.3)\end{tabular} & \begin{tabular}[c]{@{}c@{}}$\lambda$ (7000)\\ $\mu$(20), $\rho$(0.2)\end{tabular} & \begin{tabular}[c]{@{}c@{}}$\lambda$ (1000)\\  $\mu$(7), $\rho$(0.5)\end{tabular} & \begin{tabular}[c]{@{}c@{}}$\lambda$ (1.5)\\ $\mu$(0.5), $\rho$(0.1)\end{tabular} & \begin{tabular}[c]{@{}c@{}}$\lambda$ (400000)\\ $\mu$(40), $\rho$(0.4)\end{tabular} \\  
EWC                            & $\lambda$ (10000)                                                                & $\lambda$ (25000)                                                                 & $\lambda$ (500000)                                                                & $\lambda$ (40)                                                                    & $\lambda$ (1000)                                                                    \\
SI                             & c (1.0)                                                                          & c (0.7)                                                                           & c (0.85)                                                                          & c (0.75)                                                                          & -                                                                                   \\
RWALK                          & $\lambda$ (8)                                                                    & $\lambda$ (6)                                                                     & $\lambda$ (70)                                                                    & $\lambda$ (50)                                                                    & -                                                                                   \\
MAS                            & $\lambda$ (4)                                                                    & $\lambda$ (1)                                                                     & $\lambda$ (7)                                                                     & $\lambda$ (0.6)                                                                   & $\lambda$ (0.1)                                                                     \\ 
HAT                            & c (2.5), smax(400)                                                               & c (0.1), smax(400)                                                                & c (2.5), smax(400)                                                                & -                                                                                 & -                                                                                   \\ \hline
\end{tabular}}
\label{table:hyperparameter_supervised}
\end{table}

\subsubsection{Details on network architectures}
The details on network architectures for CIFAR-100, CIFAR-10/100 and Omniglot are in Table \ref{table:CIFAR_network} and \ref{table:Omniglot_network}. Since the number of classes for each task is different in Omniglot, we denoted the classes of $i$th task as $C_i$. For CUB200, we use the AlexNet architecture from PyTorch official models. \textcolor{magenta}{\footnote{\textcolor{magenta}{https://github.com/pytorch/vision/blob/master/torchvision/models/alexnet.py}}}. For the sequence of 8 different datasets, we use the model of which the size of kernel is changed to $3\times3$ and the rest is the same as AlexNet.

\begin{table}[h]
\small
\centering
\caption{Network architecture for CIFAR-100  and CIFAR-10/100}
\begin{tabular}{cccccc}
\hline
Layer                           & Channel & Kernel          & Stride & Padding & Dropout \\ \hline
32$\times$32 input              & 3     &                   &      &      &         \\ 
Conv 1                          & 32    & 3$\times$3 & 1    & 1    &                \\ 
Conv 2                          & 32    & 3$\times$3 & 1    & 1    &                \\ 
MaxPool                         &       &            & 2    & 0    & 0.25    \\ 
Conv 3                          & 64    & 3$\times$3 & 1    & 1    &                \\ 
Conv 4                          & 64    & 3$\times$3 & 1    & 1    &                \\ 
MaxPool                         &       &            & 2    & 0    & 0.25    \\ 
Conv 5                          & 128   & 3$\times$3 & 1    & 1    &                \\ 
Conv 6                          & 128   & 3$\times$3 & 1    & 1    &                \\ 
MaxPool                         &       &            & 2    & 1    & 0.25    \\ 
Dense 1                         & 256   &                   &      &      &         \\ \hline
Task 1  :  Dense 10             &       &                   &      &      &         \\
 $\cdot\cdot\cdot$                    &       &             &      &      &         \\ 
Task $i$  : Dense 10 &       &             &      &      &         \\ \hline
\end{tabular}
\label{table:CIFAR_network}
\end{table}

\begin{table}[h]
\small
\centering
\caption{Network architecture for Omniglot}
\begin{tabular}{cccccc}
\hline
Layer                   & Channel & Kernel        & Stride & Padding & Dropout \\ \hline
28$\times$28 input      & 1     &             &      &      &         \\ 
Conv 1                  & 64    & 3$\times$3  & 1    & 0    &         \\ 
Conv 2                  & 64    & 3$\times$3  & 1    & 0    &         \\ 
MaxPool                 &       &             & 2    & 0    & 0       \\ 
Conv 3                  & 64    & 3$\times$3  & 1    & 0    &         \\ 
Conv 4                  & 64    & 3$\times$3  & 1    & 0    &         \\ 
MaxPool                 &       &             & 2    & 0    & 0       \\ \hline
Task 1  :  Dense $C_1$  &       &             &      &      &         \\
 $\cdot\cdot\cdot$                    &       &             &      &      &         \\ 
Task $i$  : Dense $C_i$ &       &             &      &      &         \\ \hline
\end{tabular}
\label{table:Omniglot_network}
\end{table}

\subsubsection{Result tables}
\begin{table}[h]
\caption{Average accuracy(\%) and standard deviation for 5 random seeds} 
\resizebox{1.0\textwidth}{!}{\begin{tabular}{c|c|c|c|c|c|c}
\hline 
             & AGS-CL      & EWC         & SI          & RWLAK       & MAS         & HAT         \\ \hline \hline
CIFAR-100    & \textbf{64.1} ($\pm$1.7) & 60.2 ($\pm$1.1) & 60.3 ($\pm$1.3) & 58.1 ($\pm$1.7) & 61.5 ($\pm$0.9) & 59.2 ($\pm$0.7) \\ \hline
CIFAR-10/100 & \textbf{76.1} ($\pm$0.4) & 70.0 ($\pm$0.3) & 71.5 ($\pm$0.5) & 69.6 ($\pm$1.1) & 72.1 ($\pm$0.7) & 59.8 ($\pm$1.6) \\ \hline
Omniglot     & \textbf{82.8} ($\pm$1.8) & 76.0 ($\pm$20.2) & 54.9 ($\pm$16.2) & 71.0 ($\pm$5.6) & 81.4 ($\pm$2.1) & 5.5 ($\pm$11.1) \\ \hline
CUB200       & \textbf{81.9} ($\pm$0.7) & 80.5 ($\pm$1.2) & 80.4 ($\pm$0.8) & 81.0 ($\pm$1.3) & 79.6 ($\pm$1.0) &   -   \\ \hline
\begin{tabular}[c]{@{}c@{}}Sequence of \\ 8 different datasets\end{tabular} & \textbf{57.7} ($\pm$0.7) & 52.2 ($\pm$2.9) & - & - & 41.5 ($\pm$4.2) &   -   \\ \hline

\end{tabular}}
\label{table:Supervised detail}
\end{table}

Table \ref{table:Supervised detail}\ shows the detailed results used to generate (Figure 4, manuscript). The number in the paranthesis with $\pm$ sign stands for the standard deviation of the accuracy obtained from 5 independent runs with different random seeds. 

\subsection{Reinforcement learning}

\subsubsection{Details on network architectures}

For training Atari 8 tasks, we used the same architecture which was proposed in \cite{(atari)mnih2013playing}. However, to secure the model capacity for training 8 tasks well enough, we implemented each layer that has four times more filters than the original architecture. Figure \ref{table:Atari_network} shows the details of our model.

\begin{table}[h]
\small
\centering
\caption{Network architecture for Atari}
\begin{tabular}{cccccc}
\hline
Layer                   & Channel & Kernel        & Stride & Padding & Dropout \\ \hline
84$\times$84 input      & 4     &             &      &      &         \\ 
Conv 1                  & 32$\times$4    & 8$\times$8  & 4    & 0    &         \\ 
ReLU                    &       &             &      &      &         \\ 
Conv 2                  & 32$\times$4    & 4$\times$4  & 2    & 0    &         \\ 
ReLU                    &       &             &      &      &         \\ 
Conv 2                  & 64$\times$4    & 3$\times$3  & 1    & 0    &         \\ 
ReLU                    &       &             &      &      &         \\ 
Flatten                    &       &             &      &      &         \\
Linear1                 &       32$\times$4$\times$7$\times$7&             &      &      &         \\  \hline
Task 1  :  Dense $C_1$  &       &             &      &      &         \\
 $\cdot\cdot\cdot$                    &       &             &      &      &         \\ 
Task $i$  : Dense $C_i$ &       &             &      &      &         \\ \hline
\end{tabular}
\label{table:Atari_network}
\end{table}

\subsubsection{Hyperparameters of PPO}

We used PPO \cite{(PPO)SchulmanWlskiKlimov} as an algorithm for training Atari 8 tasks. Figure \ref{table:ppo_hyperparameters} shows hyperparameters that we used for 8 tasks, and these hyperparameters are equally applied to each baseline.  We evaluate each method every 40 updates, \textit{i.e.} we have 30 evaluation results during training each task. 
We trained the model using Adam optimizer with the initial learning rate of 0.0003 and the other hyperparameters are same as  \cite{(PPO)SchulmanWlskiKlimov}.

\begin{table}[ht]
\small
\centering
\caption{Details on hyperparameters of PPO.
}\vspace{-0.0in}\label{table:parameters}
\resizebox{0.4\linewidth}{!}{
\begin{tabular}{|c|c|}

\hline

Hyperparameters                  & Value \\ \hline \hline
\# of steps of each task          & $10^7$ \\ \hline
\# of processes                   & 128   \\ \hline
\# of steps per iteration         & 64    \\ \hline
PPO epochs                       & 10    \\ \hline
entropy coefficient              & 0     \\ \hline
value loss coefficient           & 0.5   \\ \hline
$\gamma$ for accumulated rewards & 0.99  \\ \hline
$\lambda$ for GAE                     & 0.95  \\ \hline
mini-batch size                  & 64    \\ \hline
\end{tabular}
}
\label{table:ppo_hyperparameters}
\end{table}

\newpage

\subsubsection{Detailed experimental results with $\mu = 0.1$}\label{detailed_mu01}

\begin{figure}[ht]
    \centering
    \subfigure{
    \includegraphics[width=0.90\textwidth]{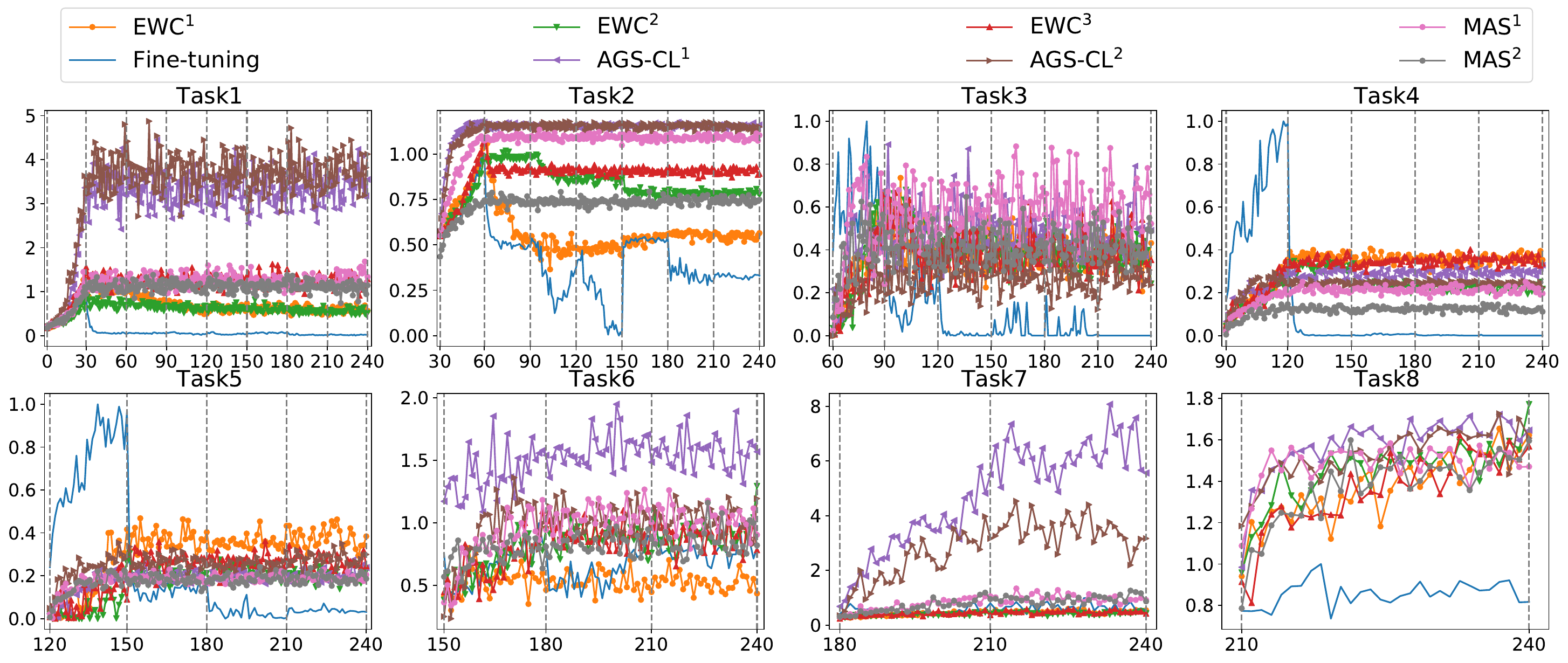}}
    \vspace{-.1in}
    \caption{Reinforcement learning results. $\lambda=\{1,2.5,10\}\times10^4$ for EWC$^{1,2,3}$, $\lambda=\{1,10\}$ for MAS$^{1,2}$, and $\mu= 0.1$, $\lambda=\{1,10\}\times 10^2$ for AGS-CL$^{1,2}$ were used, respectively.  }\label{fig:rl_mu01}
\end{figure}

Figure \ref{fig:rl_mu01} shows detailed rewards during training each task. From this figure, we can clearly observe that AGS-CL outperforms EWC for Task 1, 2 and 7 significantly. Especially, for Task 7, AGS-CL showed higher rewards than Fine-tuning, which means it achieves significantly higher plasticity. We also note that AGS-CL has higher stability than other baselines for all $\lambda$.

\begin{figure}[ht]
    \centering
    \subfigure{
    \includegraphics[width=0.5\textwidth]{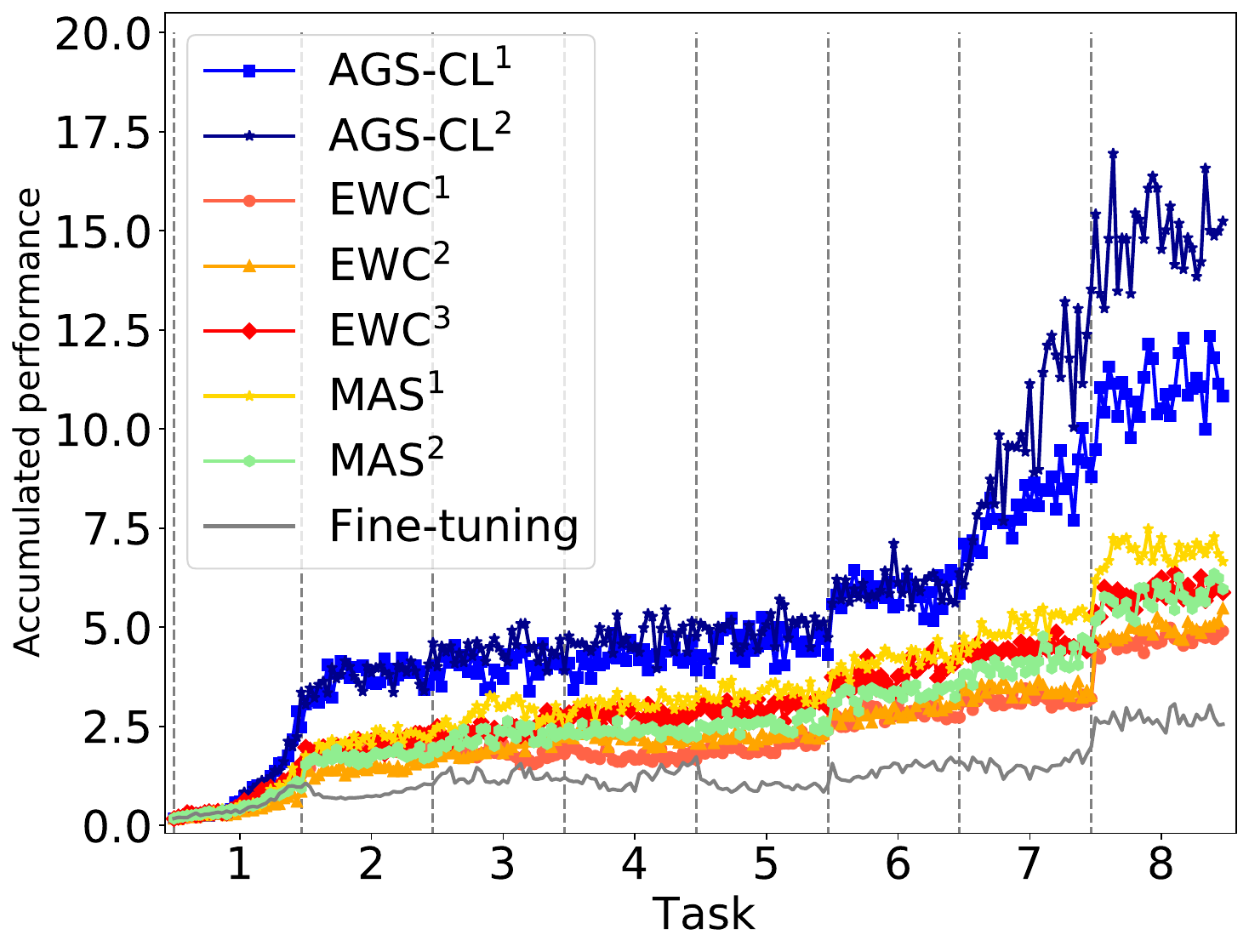}}
    \vspace{-.1in}
    \caption{Normalized accumulated rewards. $\lambda=\{1,2.5,10\}\times10^4$ for EWC$^{1,2,3}$, $\lambda=\{1,10\}$ for MAS$^{1,2}$, and $\mu= 0.125$, $\lambda=\{1,10\}\times 10^2$ for AGS-CL$^{1,2}$ were used, respectively.   }\label{fig:rl_cumulative_mu0125}
\end{figure}

\begin{figure}[ht!]
    \centering
    \subfigure{
    \includegraphics[width=0.8\textwidth]{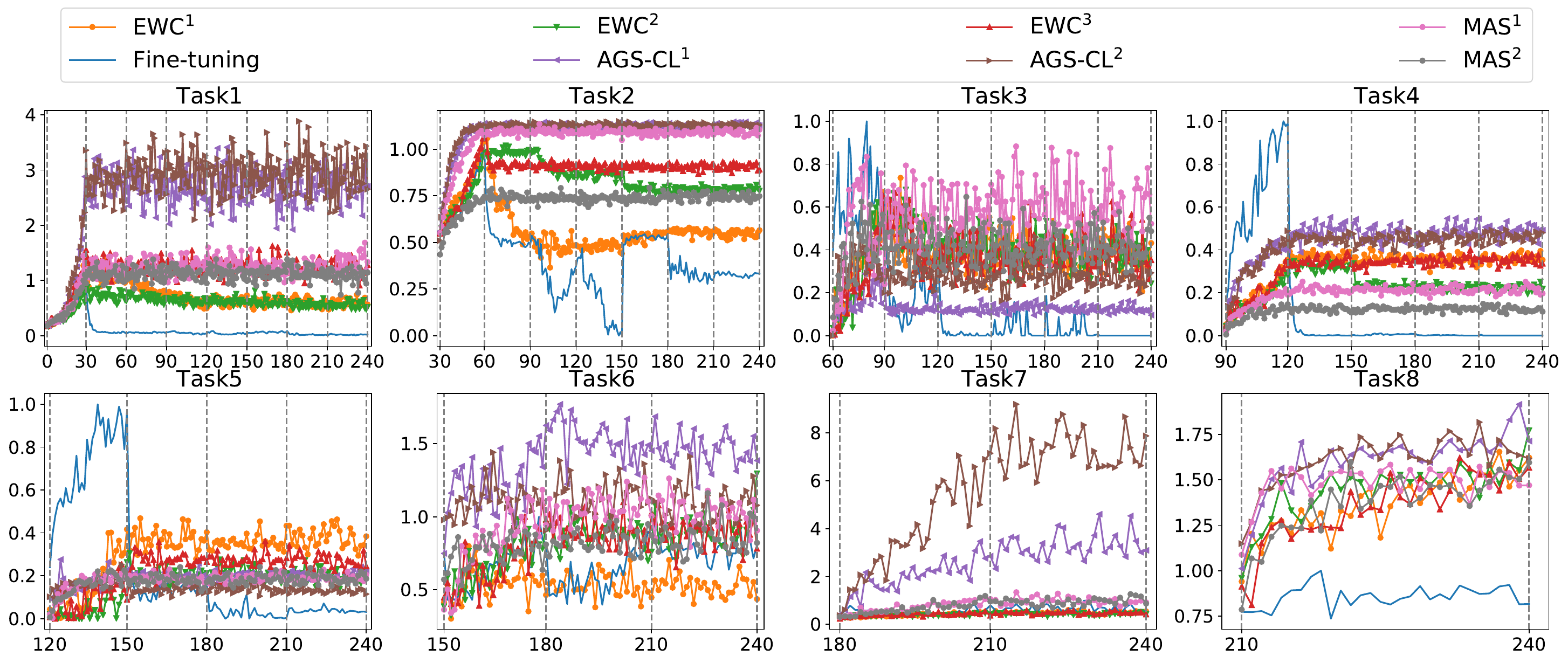}}
    \vspace{-.1in}
    \caption{Reinforcement learning results. $\lambda=\{1,2.5,10\}\times10^4$ for EWC$^{1,2,3}$, $\lambda=\{1,10\}$ for MAS$^{1,2}$, and $\mu= 0.125$, $\lambda=\{1,10\}\times 10^2$ for AGS-CL$^{1,2}$ were used, respectively.  }\label{fig:rl_mu0125}
\end{figure}

\subsubsection{Additional experimental results with $\mu = 0.125$}

To show the other result with a different $\mu$, we selected $\mu = 0.125$ and experimented in Atari 8 tasks. From Figure \ref{fig:rl_cumulative_mu0125}, we observed that AGS-CL also achieves the highest reward , which is proposed in the manuscript, using $\mu = 0.1$ if we set an appropriate $\lambda$ for AGS-CL. Figure \ref{fig:rl_mu0125} shows detailed experimental results with $\mu = 0.125$. There is a little difference with the reward of each task in Figure \ref{fig:rl_mu01} but we observed that AGS-CL shows similar advantages which we already mentioned in Section \ref{detailed_mu01}.

\bibliographystyle{plain}
\bibliography{bibfile}

%% file: intro.tex



\section{Introduction}

Continual learning, also referred to as lifelong learning, is a long standing open problem in machine learning, in which the training data is given sequentially in a form divided into the groups of tasks. The goal of continual learning is to overcome the fundamental trade-off: the \textit{stability-plasticity dilemma} \cite{(spdilemma)carpenter87, (spdilemma)mermillod13}, \ie, if the model focuses too much on the stability, it suffers from poor forward transfer to the new task, and if it focuses too much on the plasticity, it suffers from the catastrophic forgetting of past tasks.
To address this dilemma, a comprehensive study for neural network-based continual learning was conducted broadly under the following categories: regularization-based \cite{(LwF)LiHoiem16, (EWC)KirkPascRabi17, (SI)ZenkePooleGang17,(VCL)NguLiBuiTurner18,(UCL)ahn2019uncertainty,aljundi2018selfless}, dynamic architecture-based \cite{(PNN)RusuRabiDesjSoyeKirk2016, (DEN)YoonYangLeeHwang18,(Npruning)golkar2019continual}, and replay memory-based \cite{(icarl)rebuffi17, (GEM)LopezRanzato17,(DGR)ShinLeeKimKim17, (Fearnet)kemker17} methods. 

In this paper, we focus on the regularization-based methods, since they pursue to use the fixed-capacity neural network model as efficiently as possible, which may potentially allow them to be combined with other approaches. These methods typically identify important learned \textit{weights} for previous tasks and heavily penalize their deviations while learning new tasks. 
They have a natural connection with a separate line of research, the model compression of neural networks \cite{(PruningEfficient)Li2016,(LearningEfficientConvNet)Liu2017,(DiscriminationPruning)Zhuang2018}. Namely, in order to obtain a compact model, typical model compression methods measure the importance of each node or weight in a given neural network and prune the unimportant ones, hence, share the similar principle with the regularization-based continual learning schemes.  
 Several representative model compression methods \cite{wen2016learning,alvarez2016learning,yoon2017combined,scardapane2017group} used the group Lasso-like penalties, which define the incoming or outgoing weights to a \textit{node} as groups and achieve structured sparsity within a neural network. Such focus on the node-level importance could lead to a more efficient representation of the model and achieved better compression than focusing on the weight-wise importance.

Inspired by such connection, we propose a new regularization-based continual learning method, dubbed as Adaptive Group Sparsity based Continual Learning (AGS-CL), that can adaptively control the plasticity and stability of a neural network learner by using two \textit{node-wise} group sparsity-based penalties as regularization terms. Namely, our first term, which is equivalent to the ordinary group Lasso penalty, assigns and learns new important nodes when learning a new task while maintaining the structured sparsity (\ie, controls plasticity), whereas the second term, which is a group sparsity penalty imposed on the \textit{drifts} of the important node parameters, prevents the forgetting of the previously learned important nodes via freezing the incoming weights to the nodes (\ie, controls stability). The two terms are selectively applied to each node based on the \textit{adaptive} regularization parameter that represents the importance of each node, which is updated after learning each new task. For learning, we utilize the proximal gradient descent (PGD) \cite{parikh2014proximal} such that the \textit{exact} sparsity and freezing of the nodes can be elegantly attained, without any additional threshold to tune. 
Moreover, as a critical detail, we re-initialize the weights associated with the \textit{unimportant} nodes after learning each task, such that the negative transfer can be prevented and plasticity can be maximized.

As a result, we convincingly show our AGS-CL efficiently mitigates the catastrophic forgetting while continuously learning new tasks, throughout extensive experiments on several benchmarks in \textit{both} supervised and reinforcement learning. Our experimental contributions are multifold. First, we show that AGS-CL significantly outperforms strong state-of-the-art baselines \cite{(EWC)KirkPascRabi17,(SI)ZenkePooleGang17,(MAS)aljundi2018memory,(Rwalk)chaudhry2018riemannian} on \textit{all} of benchmark datasets we tested. Second, we give a detailed analysis on the stability-plasticity trade-off of our model, by utilizing additional metrics beyond average accuracy. Third, we identify AGS-CL uses \textit{orders of magnitude} less additional memory than the baselines to store the regularization parameters, thanks to only maintaining the node-wise regularization parameters. Such compact memory usage is a nice by-product and enables applying our method to much larger networks, which typically is necessary for applications with large-scale datasets. Finally, we stress that our RL results on Atari games are for the \textit{pure} continual learning setting, in which past tasks cannot be learned again, in contrast to other works \cite{(EWC)KirkPascRabi17,(ProgressCompress)SchwarzLuketinaHadsell18} that allow the agents to learn multiple tasks in a \textit{recurring} fashion. 

\noindent\textbf{Related work}\ \ 
 Diverse approaches for neural networks based continual learning have been proposed, as exhaustively surveyed in \cite{(Review)PariKemPartKananWerm18}.
 Unlike the typical weight-wise regularization-based methods, \textit{e.g.},  \cite{(EWC)KirkPascRabi17,(SI)ZenkePooleGang17,(Rwalk)chaudhry2018riemannian,(MAS)aljundi2018memory,(VCL)NguLiBuiTurner18}, several other works considered the node-wise importance, similarly as ours, as well, but had some limitations. For instance, \cite{(UCL)ahn2019uncertainty} considered node-importance in the context of Bayesian neural network and variational inference, but their method had to work with several heuristic-based losses and cannot be applied to non-Bayesian pre-trained models. \cite{SerraSurisMironKarat2018(HAT)} utilized a node-wise hard attention mechanism per layer to freeze the important nodes, but they required to know the total number of tasks to be learned in advance and had to implement a subtle annealing heuristic for attention. \cite{aljundi2018selfless} devised additional regularization term for promoting node-wise sparsity to boost the performance of the weight-wise regularization based methods, but the scheme still had to store the weight-wise regularization parameters. \cite{(Npruning)golkar2019continual} developed a notion of active and inactive nodes and implemented pruning and freezing schemes, which are similar to ours, but they required several hyperparameters and involved cumbersome, non-adaptive threshold tuning steps that require a separate validation set. In contrast, our AGS-CL employs more principled loss function and optimization routine, unlike \cite{(UCL)ahn2019uncertainty,SerraSurisMironKarat2018(HAT),(Npruning)golkar2019continual}, stores only node-wise regularization parameters, unlike \cite{aljundi2018selfless}, and automatically determines which nodes to prune or freeze, unlike \cite{(Npruning)golkar2019continual}.

The group Lasso \cite{yuan2006model} regularization, which was favorably used in model compression \cite{alvarez2016learning, wen2016learning, yoon2017combined, scardapane2017group}, has been also adopted for continual learning in \cite{(DEN)YoonYangLeeHwang18}. However, they considered a setting in which model capacity can grow as the learning continues, which is different from our focus, and their method involved many hyperparameters and multiple re-training steps, which make it hard to apply in practice. Moreover, there was no mechanism to freeze the model at the group level in \cite{(DEN)YoonYangLeeHwang18}.

%% file: main.tex
\section{Motivation}\label{subsec:motivation}

\begin{wrapfigure}{r}{0.4\textwidth}
    \centering
    \includegraphics[width=0.39\textwidth]{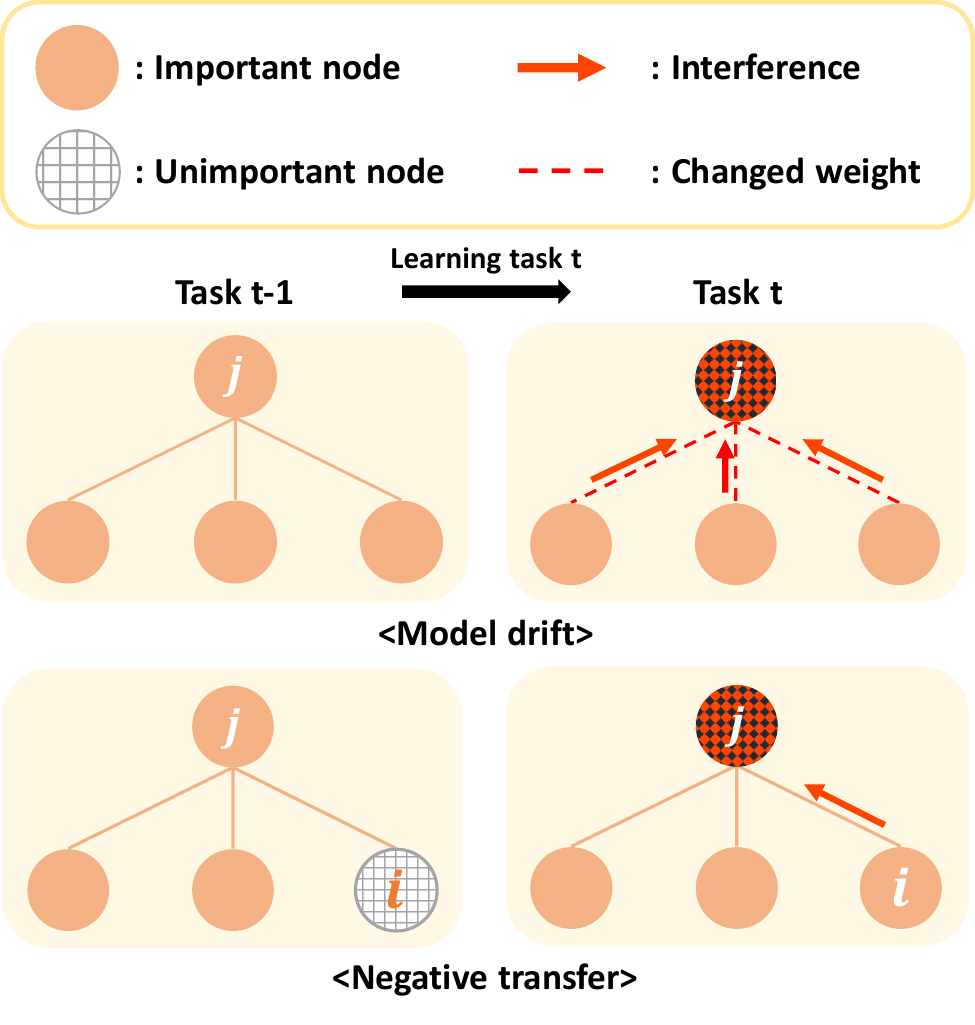}    
    \caption{Two sources of catastrophic forgetting: model drift (top) and negative transfer (bottom). 
    }\label{fig:motivation}
    \vspace{-.2in}
\end{wrapfigure}

Here, we give the main motivation for our algorithm. 
We start from the intuition that the node in a neural network is the basic unit for representing the learned information from a task, and the catastrophic forgetting occurs when the information flowing to the important nodes changes as the learning continues with new tasks.
Namely, assume the important nodes for task $t-1$ are identified. Then, we argue that there are two sources for the catastrophic forgetting: \textit{model drift} and \textit{negative transfer} as shown in Figure \ref{fig:motivation}. The model drift corresponds to the case in which the incoming weights of an important node (node $j$ in Figure \ref{fig:motivation}) gets changed when learning a new task $t$.
 In this case, the representation of node $j$ for task $t-1$ can alter, hence, the performance for task $t-1$ can drastically degrade. The red arrows and dotted lines in the figure exemplify such model drift for node $j$. On the other hand, the negative transfer happens when the representation of an unimportant node for task $t-1$ in the lower layer (node $i$ in Figure \ref{fig:motivation}) changes during learning a new task $t$. Namely, even when there is no model drift, if node $i$ becomes important for task $t$, then such change of representation will bring an interfering effect for node $j$ when carrying out task $t-1$. The color change of $i$ and red arrow in the figure shows such negative transfer from the future tasks. \vspace{.02in}
In order to address above two key sources of catastrophic forgetting, we believe an effective continual learning algorithm should essentially carry out the followings:
\begin{compactitem}
\item \emph{Freeze} important nodes: Once a node has been identified as important, its \textit{incoming} weights should be frozen while learning future tasks, hence, the model drift can be prevented. 
\item \textit{Nullify} transfer from unimportant nodes: Once a node has been identified as unimportant, its \textit{outgoing} weights should be fixed to $0$ (\textit{i.e.}, pruned), hence, the negative transfer from the node to the upper layers can be eliminated. 
\end{compactitem}
We note most of the state-of-the-art regularization-based methods aim to approximate the first item via regularizing the important weights, while largely neglecting the second item. One exception is \cite{(Npruning)golkar2019continual}, but as mentioned in related work, their method required multiple heuristics to determine unimportant nodes and prune the outgoing weights of the unimportant nodes. 
Our proposed AGS-CL, on the other hand, automatically determines the important and unimportant nodes as the learning continues, freezes the incoming weights for the important ones, and nullifies the outgoing weights for the unimportant ones, all via selectively applying two group sparsity based penalties based on the adaptive regularization parameter defined for each node. Furthermore, to maximize the plasticity, the random initialization of the \textit{incoming} weights of unimportant nodes are implemented as well, and we elaborate each step more in details in the next section.

\vspace{-.1in}

\section{Adaptive Group Sparsity based Continual Learning (AGS-CL)}\label{sec:ags-cl}
\vspace{-.02in}

\subsection{Notations}
\vspace{-.02in}

We denote $\ell\in\{1\ldots,L\}$ as a layer of a neural network model that has $N_\ell$ nodes, and let $n_\ell\in\{1,\ldots,N_{\ell}\}$ be a node in that layer. For the convolutional neural networks (CNN), a node stands for a convolution filter (or channel). Moreover, $\btheta_{n_\ell}$ denotes the vector of the (incoming) weight parameters for the $n_\ell$-th node. Hence, $\theta_{n_\ell,i}$ stands for the weight that connects the $i$-th node (or channel) in layer $\ell-1$ with the node $n_\ell$. Moreover, $\mathcal{G}\triangleq\{n_\ell:1\leq n_\ell\leq N_\ell, 1\leq \ell\leq L\}$ is the set of all the nodes in the neural network, and $\btheta=\{\btheta_{n_{\ell}}\}_{n_\ell\in\mathcal{G}}$ denotes the entire parameter vector of the network. We assume ReLU is always used as the activation function for all layers. 
We denote $\mathcal{D}_t$ as the training dataset for task $t\in\{1,\ldots,\mathcal{T}\}$, and we assume the task boundaries are given to the learner.




\vspace{-.02in}
\subsection{Loss function}\label{subsec:lossf}
\vspace{-.02in}

Before describing the loss function for task $t$, we first introduce the adaptive regularization parameter $\Omega_{n_\ell}^{t-1}\geq0$ for each node $n_\ell\in\mathcal{G}$, of which magnitude indicates how important the node is for carrying out the tasks up to $t-1$. Namely, large $\Omega_{n_\ell}^{t-1}$ indicates that the node $n_\ell$ has been identified and learned as important, and $\Omega_{n_\ell}^{t-1}=0$ denotes the node $n_\ell$ was not important for learning tasks up to $t-1$. 
The exact definition and update mechanism for $\Omega_{n_\ell}^{t-1}$ are given in Section \ref{subsubsec:re_init}, but for now, we assume such parameter is given when learning a new task $t$. 

Given such $\{\Omega_{n_\ell}^{t-1}\}_{n_\ell\in\Gcal}$, we define a set of unimportant nodes as
\begin{align}
    \Gcalz^{t-1}\triangleq\{n_\ell:\olm=0\}\subseteq\Gcal,\label{eq:g_zero}
\end{align}
and with the training data $\Dcal_t$, our loss function for learning task $t$ is defined as 
\begin{align}
\mathcal{L}_t(\bm\theta) = 
\mathcal{L}_{\text{TS},t}(\btheta)
+ \underbrace{\mu\sum_{n_\ell\in\mathcal{G}_0^{t-1}}\|\bthetanl\|_2}_{(a)}
+ \underbrace{\lambda \sum_{n_\ell\in\mathcal{G}\backslash \mathcal{G}_0^{t-1}}\olm\|\bthetanl-\hatbtheta\|_2}_{(b)}.\label{eq:loss_function}
\end{align}
In (\ref{eq:loss_function}), $\mathcal{L}_{\text{TS},t}(\btheta)$ stands for the ordinary task-specific loss on $\mathcal{D}_t$ (\textit{e.g.}, cross-entropy for supervised learning), and the terms (\textit{a}) and (\textit{b}) are the group sparsity-based regularization terms, in which 
$\hatbtheta$ is the learned parameter vector for node $n_\ell$ up to task $t-1$, and $\mu,\lambda\geq 0$ are the hyperparameters that set the trade-offs among the penalty terms. 
Notice that our loss function \emph{selectively} employs the regularization terms based on the value of $\Omega_{n_\ell}^{t-1}$. Namely, for the unimportant nodes in $\Gcalz^{t-1}$, we apply the group Lasso penalty (term (\textit{a})) as in \cite{alvarez2016learning}, and for the important nodes in $\Gcal\backslash\Gcalz^{t-1}$, we apply the group-sparsity based deviation penalty (term (\textit{b})) that adaptively penalizes (similarly as in \cite{Zou06,(GroupLasso)wang2008note}) the deviation of $\btheta_{n_{\ell}}$ from $\hat{\btheta}_{n_{\ell}}^{(t-1)}$ depending on the magnitude of $\olm>0$. 

We elaborate that term (\textit{a}) controls the \textit{plasticity} of the model when learning new tasks, whereas term (\textit{b}) is in charge of achieving the \textit{stability} via preventing the model drift mentioned in Section \ref{subsec:motivation}. 
Namely, term (\textit{a}) automatically identifies the active learners for the new task $t$ among the unimportant nodes so far and sparsifies the rest of the nodes such that they can be allocated for learning future tasks. On the other hand, the term (\textit{b}) enforces to freeze a node (\textit{i.e.}, prevent model drift) if it has been identified to be important enough, \ie, it has large $\olm$ value. Note due to the property of the group-norm penalties, the sparsification and freezing resulting from applying the two regularization terms can be \textit{exact} when appropriate optimization method is used, as described in the next subsection.

\begin{figure*}[t]
    \centering
    \includegraphics[width=1.0\textwidth]{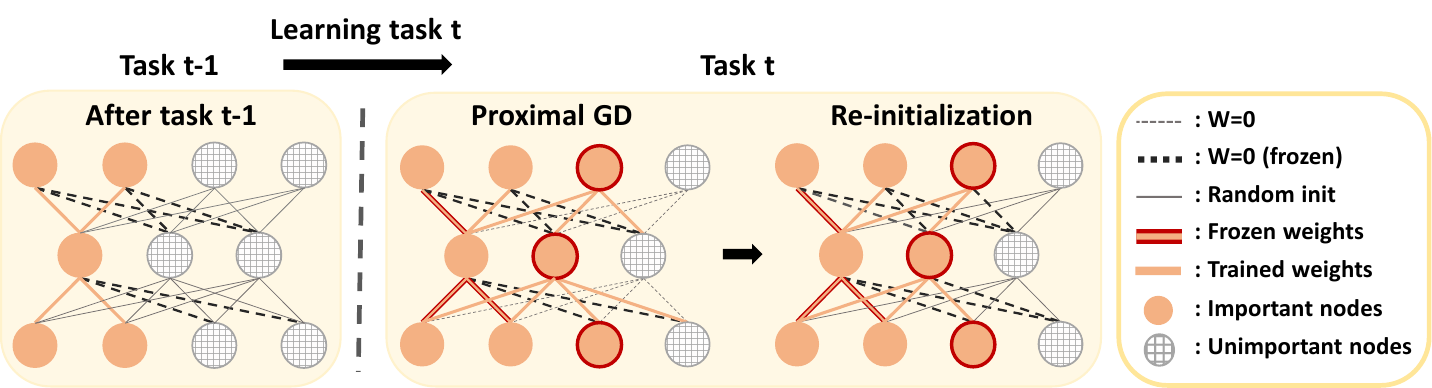}
    \caption{Summary of AGS-CL. During learning a new task $t$, the PGD step using term (\textit{a}) in (\ref{eq:loss_function}) identifies the \textit{new} important nodes (orange, solid nodes with red boundaries) and remaining unimportant nodes (gray, checkered nodes). The incoming weights connected to the \textit{sufficiently} important nodes up to $t-1$ (vanilla orange, solid nodes) are frozen at $t$ (orange lines with red boundaries) due to the PGD step with term (\textit{b}) in (\ref{eq:loss_function}). The re-initialization step then fixes the outgoing weights of unimportant nodes to zero (black, bold dotted lines) and randomly initializes the incoming weights of unimportant nodes
        (gray, solid thin weights).
    }\label{fig:gs_process}
    \vspace{-.1in}
\end{figure*}

\subsection{Learning with proximal gradient descent}\label{sec:pgd}

While directly minimizing $\mathcal{L}_t(\btheta)$ can be done via applying vanilla SGD-variant optimizers, \eg, Adam \cite{KinBa15}, we employ the proximal gradient descent (PGD) method \cite[Section 4.2]{parikh2014proximal}.
To that end, we first denote the proximal operator as
\begin{align}
\bprox_{\alpha f}(\bm v)=\arg\min_{\btheta}\Big(f(\btheta)+\frac{1}{2\alpha}\|\btheta - \bm v\|_2^2\Big)\label{eq:prox}
\end{align}
for a  scalar $\alpha>0$ and a convex function $f$. 
Then, by simply denoting (\ref{eq:loss_function}) as $\mathcal{L}_t(\btheta) = \mathcal{L}_{\text{TS},t}(\btheta) + \mathcal{L}_{\text{Reg},t}(\btheta)$, in which $\mathcal{L}_{\text{Reg},t}(\btheta)$ is the \textit{convex} regularization term that combines term (\textit{a}) and term (\textit{b}) in (\ref{eq:loss_function}), the PGD with learning rate $\alpha$ 
iteratively minimizes (\ref{eq:loss_function}) by computing the following
\begin{align}
\tilde{\btheta}^{k+1}\ :=&\ \ \btheta^k-\alpha \nabla \mathcal{L}_{\text{TS},t}(\btheta^k)\label{eq:ce_update}\\
\btheta^{k+1}\ :=&\ \ \textbf{prox}_{\alpha\mathcal{L}_{\text{Reg},t}(\btheta)} \big(\tilde{\btheta}^{k+1}\big).\label{eq:prox_gradient}
\end{align}
for $k=0,\ldots,\mathcal{K}-1$. Namely, $\btheta^k$ is the $k$-th proximal gradient update step. Namely, (\ref{eq:prox_gradient}) applies the proximal operator (\ref{eq:prox}) with $f=\mathcal{L}_{\text{Reg},t}(\btheta)$ on the gradient update of $\btheta^k$ using $\nabla \mathcal{L}_{\text{TS}}(\btheta^k)$. 
Now, for deriving a succinct, concrete parameter update rule for our algorithm, we introduce the following lemma, of which proof is given in the Supplementary Material.



\begin{lemma}\label{eq:lem}
For $f(\btheta)=c\|\btheta-\btheta_0\|_2$ with $c>0$ and any fixed vector $\btheta_0$,
\begin{align}
\emph{\bprox}_{\alpha f}(\bm v) = \gamma \bm v+ (1-\gamma) \btheta_0,
\end{align}
in which 
$
\gamma = \big(1-\frac{\alpha c}{\|\btheta_0-\bm v\|_2}\big)_+
$
where $(x)_+=\max\{0,x\}$.
\end{lemma}

From (\ref{eq:prox}), we can easily see that the proximal operator can be applied to each node parameter vector $\bthetanl$, or each group, independently when carrying out (\ref{eq:prox_gradient}). Hence, by Lemma \ref{eq:lem}, we have the following closed-form proximal gradient update rules:
\begin{equation}
\label{eq:update_rule}
\bthetanl^{k+1} =
    \begin{cases}
      \big(1-\frac{\alpha\mu}{\|\tilde{\btheta}^{k+1}_{n_\ell}\|_2}\big)_+\tilde{\btheta}^{k+1}_{n_\ell} \ \ \quad\quad \text{for} \ \ \ n_\ell\in\Gcal_0^{t-1}\\
      \gamma \tilde{\btheta}^{k+1}_{n_\ell}+(1-\gamma) \hatbtheta \  \quad \text{for} \ \ \ n_\ell\in\Gcal\backslash\Gcal_0^{t-1},
    \end{cases}
\end{equation}
in which 
$
\gamma = \big(1-\frac{\alpha\lambda\olm}{\|\tilde{\btheta}^{k+1}_{n_\ell}-\hatbtheta\|_2}\big)_+.
$
Note the first rule in (\ref{eq:update_rule}) can set $\bthetanl^{k+1}=\mathbf{0}$ (\ie, sparsify) when $\|\tilde{\btheta}^{k+1}_{n_\ell}\|_2\leq\alpha\mu$ for the unimportant nodes, and the second rule can set $\bthetanl^{k+1}=\hatbtheta$ (\ie, freeze) when $\|\tilde{\btheta}^{k+1}_{n_\ell}-\hatbtheta\|_2\leq\alpha\lambda\olm$ for the important nodes. 
Thus, we can automatically achieve the \emph{exact} sparsification and freezing of the node parameters as a part of the optimization routine, \textit{without} any additional thresholds or heuristics that are otherwise required when using vanilla SGD-variants or \cite{(Npruning)golkar2019continual}. Moreover, we show in the Supplementary Material that the accurate sparsification and freezing by our PGD update is integral in achieving high accuracy by comparing with a scheme without it.
Finally, from the theory of PGD \cite{parikh2014proximal}, (\ref{eq:update_rule}) is guaranteed to converge to a local minima of $\mathcal{L}_t(\btheta)$ with appropriate $\alpha$. The converged parameters
are then denoted as $\hat{\btheta}^{(t)}=\{\hat{\btheta}_{n_\ell}^{(t)}\}_{n_\ell\in\Gcal}$.

\subsection{Updating $\olm$ and re-initialization of unimportant nodes}\label{subsubsec:re_init}

\textbf{Updating $\olm$}\ \ Now, we give the definition and the update formula of  $\{\olm\}_{n_\ell\in\Gcal}$, which reflect the importance of nodes and play a crucial role in our loss function.
\begin{wrapfigure}{r}{0.3\textwidth}
    \centering
    \includegraphics[width=0.3\textwidth]{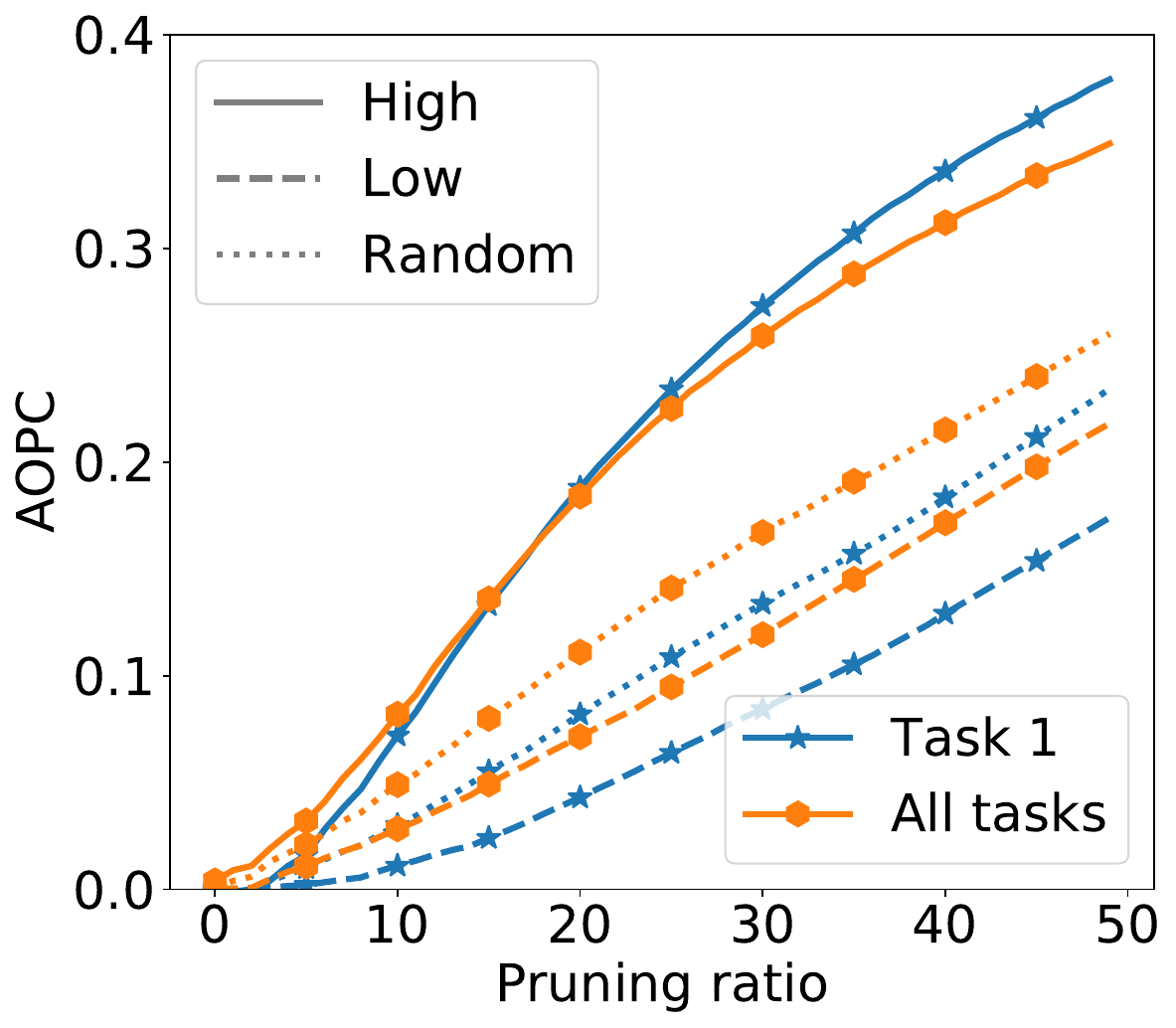}
    \vspace{-.2in}
    \caption{AOPC for $\{\Omega_{n_\ell}^{t}\}$
    }\label{fig:aopc}
    \vspace{-.27in}
\end{wrapfigure}
Initially, we set $\Omega_{n_\ell}^{0}=0$ for all $n_\ell\in\Gcal$, thus, for $t=1$, we obtain the ordinary group Lasso solution since $\Gcal_0^{0}=\Gcal$. After minimizing $\mathcal{L}_t(\btheta)$, $\Omega_{n_\ell}^{t}$ gets updated as
\begin{align}
\Omega_{n_\ell}^{t} \ := \ \ \eta \olm + \frac{1}{N_t}\sum_{m=1}^{N_t}a_{n_\ell}(\bm x_m^{(t)})\label{eq:update}
\end{align}
for all $n_\ell\in \Gcal$,
in which $a_{n_\ell}(\bm x_m^{(t)})$ is the ReLU activation value of the node $n_\ell$ when the input data is $\bm x_m^{(t)}\in\mathcal{D}_t$, and $\eta\in(0,1]$ is the hyperparameter for the exponential averaging. Hence, we regard the average activation value of $n_\ell$ for task $t$ as the \emph{importance} of the node, and it is added to $\olm$. Namely, a node remains unimportant (\textit{i.e.}, be in $\Gcal_0^t$) when either the incoming weights remain to be zero or the ReLU activations are dead for all training data points, after learning task $t$. 
Furthermore, $\eta<1$ implements exponential moving average, similarly as in \cite{(ProgressCompress)SchwarzLuketinaHadsell18}, such that the $\{\Omega_{n_\ell}^{t}\}$ values do not explode (we always used $\eta=0.9$).

One may argue whether the average ReLU activation as in (\ref{eq:update}) can be a correct measure for identifying the importance of a node. To that end, Figure \ref{fig:aopc} justifies our choice by considering  Area Over Prediction Curve (AOPC) \cite{samek2016evaluating} for $\{\Omega_{n_{\ell}}^t\}$ on CIFAR-100 \cite{(cifar)krizhevsky2009learning} tasks, which splits 100 classes into 10 tasks. 
AOPC is a widely used metric for quantitatively evaluating the neural network interpretation methods, \textit{e.g.}, \cite{grad-cam17,lrp}, and a steep increase of AOPC with respect to the pruning (or perturbing) of nodes (or input pixels) in the order of high importance values suggests the validity of an interpretation method. Figure \ref{fig:aopc} shows AOPC curves of our importance measure $\{\Omega_{n_{\ell}}^t\}$, in which the pruning of nodes is done in the order of random (dotted), highest (solid) and lowest (dashed) values after learning task 1 (blue line with star) and all tasks (orange line with circle), respectively. We clearly observe the significant gaps between the solid and dashed/dotted lines, which corroborates the validity of using average ReLU activation for $\{\Omega_{n_{\ell}}^t\}$. We note some alternatives for $\{\Omega_{n_{\ell}}^t\}$ may be also used, \textit{e.g.}, apply neural network interpretation methods, but due to the simplicity (\textit{i.e.}, only requiring forward-pass in contrast to \cite{grad-cam17,lrp} that also require backward-pass) and correctness shown in Figure \ref{fig:aopc}, we adhere to using (\ref{eq:update}) and defer to future work for comparing with other interpretation methods.

\noindent\textbf{Re-initialization:} \ Once $\{\Omega_{n_\ell}^{t}\}_{n_\ell\in\Gcal}$ are updated, we carry out two re-initialization steps on the weights that are connected to the unimportant nodes in $\Gcal_0^{t}$. 
That is, for the weights $\hat{\btheta}^{(t)}=\{\hat{\btheta}_{n_\ell}^{(t)}\}_{n_\ell\in\Gcal}$, 


\begin{compactitem} 
\item[(I.1)] \textbf{[Zero-init]}
Fix $\hat{\theta}^{(t)}_{n_\ell,i}=0\ \text{if}\ \ i\in\mathcal{G}_0^{t}$, for all future tasks after $t$.
\item[(I.2)] \textbf{[Rand-init]} Randomly initialize  $\hat{\btheta}^{(t)}_{n_\ell}$ if $n_\ell\in\mathcal{G}_0^{t}$, with probability $\rho$.

\end{compactitem}
The former fixes the \textit{outgoing} weights of an unimportant node to zero (\textit{i.e.}, prunes) for \textit{all} remaining tasks, while the latter randomly initializes (\textit{i.e.}, frees) the \emph{incoming} weight vector of an unimportant node, with probability $\rho$. 
We can see (I.1) prevents the negative transfer mentioned in Section \ref{subsec:motivation} and improves stability, since the change of the representations of unimportant nodes in $\Gcal_0^t$ happening in the future tasks will \textit{never} affect the important nodes for task $t$ in the upper layer that are connected to $i$. Moreover, we observe 
(I.2) enables some nodes in $\Gcalz^{t}$ to become \textit{active learners} for future tasks and improves plasticity, since the incoming weights of those nodes would otherwise typically not get updated due to zero gradient. $\rho\in(0,1]$ is a hyperparameter that controls the capacity of the network for learning new tasks, and $\rho\leq0.5$ typically shows good trade-off between the sparsity and used capacity of the network. 

We also emphasize that the order of the re-initialization steps is important, \ie, (I.1) is always followed by (I.2), which can be seen by observing that the outgoing weights of an unimportant node can be connected to \textit{either} important \textit{or} unimportant nodes in the upper layer. Namely, in such a case, (I.1) \textit{nullifies} and \textit{fixes} those connected to the important nodes,
but (I.2) re-utilizes those connected to the unimportant nodes so that they can become learnable again for the next task. 
This 
is also illustrated in the rightmost network figure in Figure \ref{fig:gs_process}; note the activation of the unimportant node in the second layer can be used for learning the unimportant node in the third layer for task $t+1$ (via the gray, solid thin weight between them that is randomly re-initialized). In our experimental results, we systematically show the critical effects of (I.1) and (I.2). 

Finally, we summarize our method in  Algorithm \ref{alg} and Figure \ref{fig:gs_process}.

\vspace{-.05in}
\begin{algorithm}[H]
\caption{AGS-CL algorithm}\label{alg}
\begin{algorithmic}
\REQUIRE $\{\mathcal{D}_t\}_{t=1}^{\Tcal}$: Sequential training datasets
\REQUIRE $\mu$, $\lambda$, $\rho$: Hyperparamters, $\mathcal{K}$ : Number of epochs for each task
\STATE Randomly initialize $\btheta$ and set $\Omega_{n_\ell}^{0}=0$, $\forall n_\ell\in\Gcal$.
\FOR{$t = 1, \cdots, \mathcal{T}$}
\STATE Define the loss function $\mathcal{L}_t(\btheta)$ in (\ref{eq:loss_function}).
  \FOR{$k = 0, \cdots, \mathcal{K}-1$} 
  \STATE Compute (\ref{eq:ce_update}) and (\ref{eq:prox_gradient}) together with (\ref{eq:update_rule}) to obtain $\btheta_{n_\ell}^{k}$ for each $n_\ell\in\Gcal$ \ \ \textcolor{orange}{\texttt{/*PGD updates*/}}
    
    
  \ENDFOR
  \STATE Obtain $\hat{\btheta}^{(t)}$ and update $\{\Omega_{n_\ell}^{t}\}_{n_\ell\in\Gcal}$ using (\ref{eq:update}) \quad  \textcolor{orange}{\texttt{/*Update $\{\Omega_{n_{\ell}}^t\}$*/}}
  \STATE Obtain $\Gcalz^{t}$ as in (\ref{eq:g_zero})
  \STATE Re-initialize $\hat{\btheta}^{(t)}$ using first with (I.1), then with (I.2) \quad  \textcolor{orange}{\texttt{/*Re-initializations*/}}
\ENDFOR
\end{algorithmic}
\label{alg:CL}
\end{algorithm}
\vspace{-.05in}

%% file: experiments.tex
\vspace{-.15in}
\section{Experimental Results}

\subsection{Supervised learning on vision datasets}

We evaluate the performance of AGS-CL together with the representative regularization-based methods, EWC \cite{(EWC)KirkPascRabi17}, SI \cite{(SI)ZenkePooleGang17}, RWALK \cite{(Rwalk)chaudhry2018riemannian}, MAS \cite{(MAS)aljundi2018memory}, and HAT \cite{SerraSurisMironKarat2018(HAT)}. We used multi-headed outputs for all experiments, and 5 different random seed runs (that also shuffle task sequences except for Omniglot) are averaged for all datasets. 
We tested on multiple different vision datasets and thoroughly showed the effectiveness of our method: CIFAR-10/100 \cite{(cifar)krizhevsky2009learning} was used as a standard benchmark with smaller number of tasks, Omniglot  \cite{(Omniglot)Lake11} was used to compare the performance for large number of tasks, CUB200 \cite{(cub)welinder2010caltech} was used to test on more complex, large-scale data, and the sequence of 8 different datasets, $\{$CIFAR-10 / CIFAR-100 / MNIST / SVHN / Fashion-MNIST / Traffic-Signs / FaceScrub / NotMNIST$\}$, which was proposed in \cite{SerraSurisMironKarat2018(HAT)}, was used to test the check the learning capability for different visual domains.

For all the experiments, we used convolutional neural networks (CNN) with ReLU activations, of which architectures are the followings: for CIFAR-10/100, we used 6 convolution layers followed by 2 fully connected layers, for Omniglot, we used 4 convolution layers as in  \cite{(ProgressCompress)SchwarzLuketinaHadsell18}, for CUB200, we used AlexNet \cite{(alexnet)krizhevsky2012imagenet} pre-trained on ImageNet \cite{(Imagenet)Deng09}, and for the mixture of different tasks, we used AlexNet trained from scratch. We fairly searched the hyperparameters for all baselines and report the best performance for each method. Our method was implemented with PyTorch \cite{(Pytorch)paszke2017automatic}, and Adam \cite{KinBa15} step was used as $\nabla \mathcal{L}_{\text{TS},t}(\btheta^k)$ in (\ref{eq:ce_update}), and PGD update (\ref{eq:prox_gradient}) was applied once after each epoch. More details and ablation studies on the hyperparameters (particularly for $\rho$ and PGD updates) as well as on model architectures with full hyperparameter settings are given in the Supplementary Material. 
\begin{figure*}[h]
    \centering
    \includegraphics[width=0.7\textwidth]{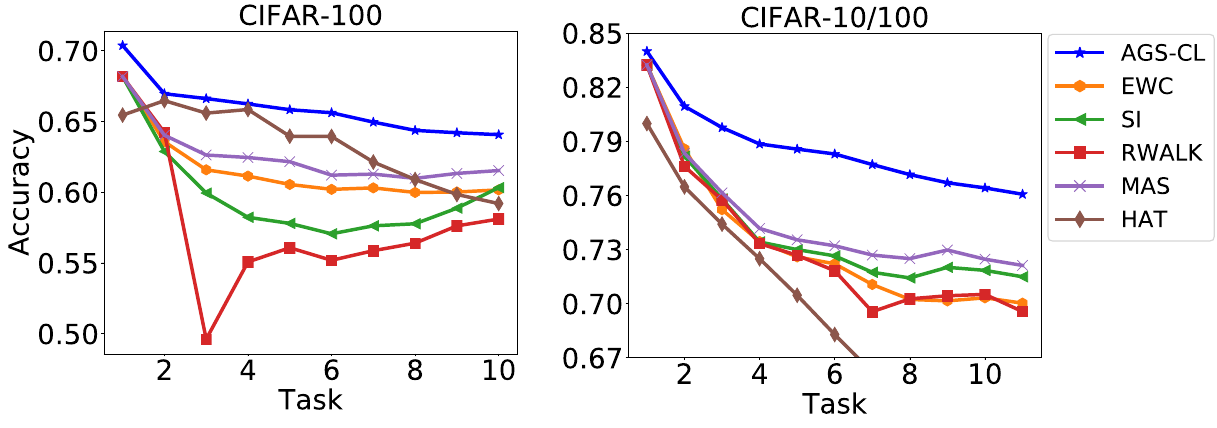}
    \vspace{-.1in}
    \caption{Average accuracy results on CIFAR-100 and CIFAR-10/100 datasets.}\label{fig:Supervised_cifar}
        \vspace{-.1in}
\end{figure*}
\begin{figure*}[h]
    \centering
    \includegraphics[width=0.9\textwidth]{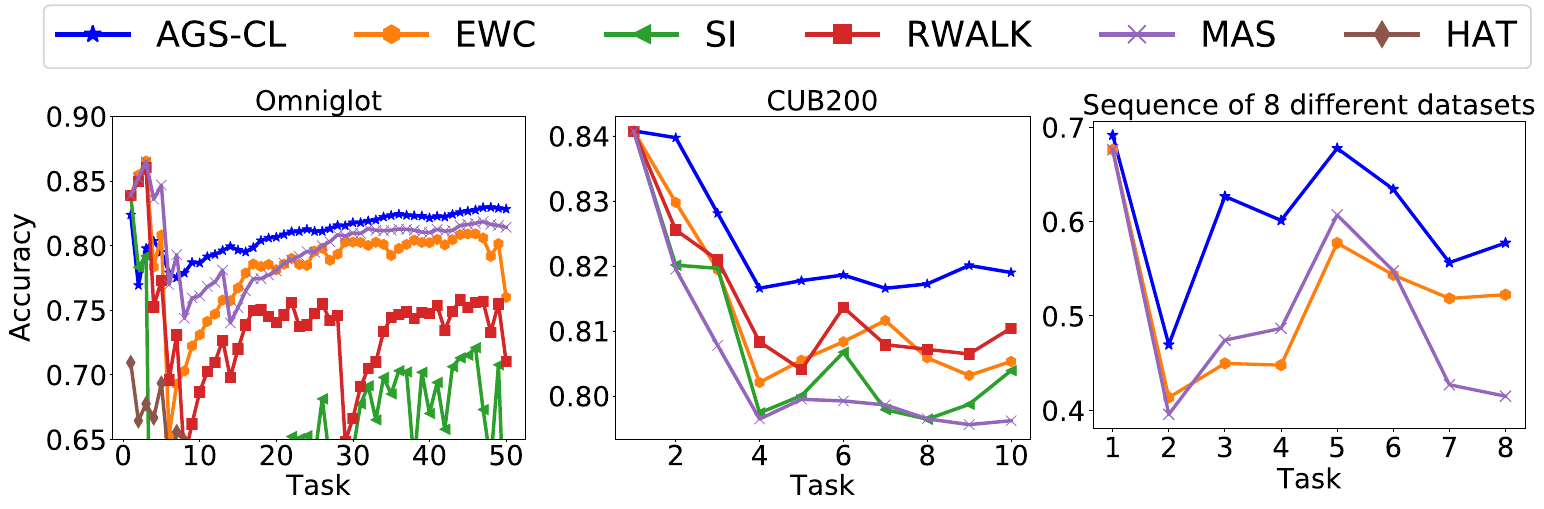}
    \vspace{-.1in}
    \caption{Average accuracy results on Omniglot, CUB200, and the sequence of 8 datasets. 
    }\label{fig:Supervised_others}
        \vspace{-.1in}
\end{figure*}

\noindent\textbf{Average accuracy}\ \ 
Figure \ref{fig:Supervised_cifar} and \ref{fig:Supervised_others} show the average accuracy result on each dataset. The first figure in Figure \ref{fig:Supervised_cifar} is on CIFAR-100, which splits 100 classes into 10 tasks with 10 classes per task, and the second is on CIFAR-10/100, which additionally uses CIFAR-10 for pre-training before learning tasks from CIFAR-100. In Figure \ref{fig:Supervised_others}, the first figure is on Omniglot, which treats each alphabet as a single task and uses all 50 alphabets, 
the second figure is on CUB200, which splits 200 classes into 10 tasks with 20 classes per task, and the third figure is on the sequence of 8 different vision datasets, which treats each dataset as a separate task. For the first and third figures, there were different numbers of classes for each task, and the total number of classes was 1600 and 293. For the sequence of 8 different datasets, we only compared with EWC and MAS since they were the two best baselines on other datasets. We can make the following observations from the results. Firstly, we clearly observe that our AGS-CL consistently dominates other baselines for all the datasets throughout most tasks. We stress that this is significant since AGS-CL uses much smaller memory to store the regularization parameters than others, as more elaborated below. Secondly, among other baselines, there is no clear winner; MAS tends to excel in the first three sets, while it is the worst for CUB200 and the 8 different vision datasets. 
Thirdly, as seen in the results for Omniglot, SI and RWALK, which are based on path integral of gradient vector field, had large performance variance for larger number of tasks.

\begin{figure}[ht]
    \centering
     \subfigure[Size of regularization parameters]{\label{fig:Parameters}
    \includegraphics[width=0.32\columnwidth]{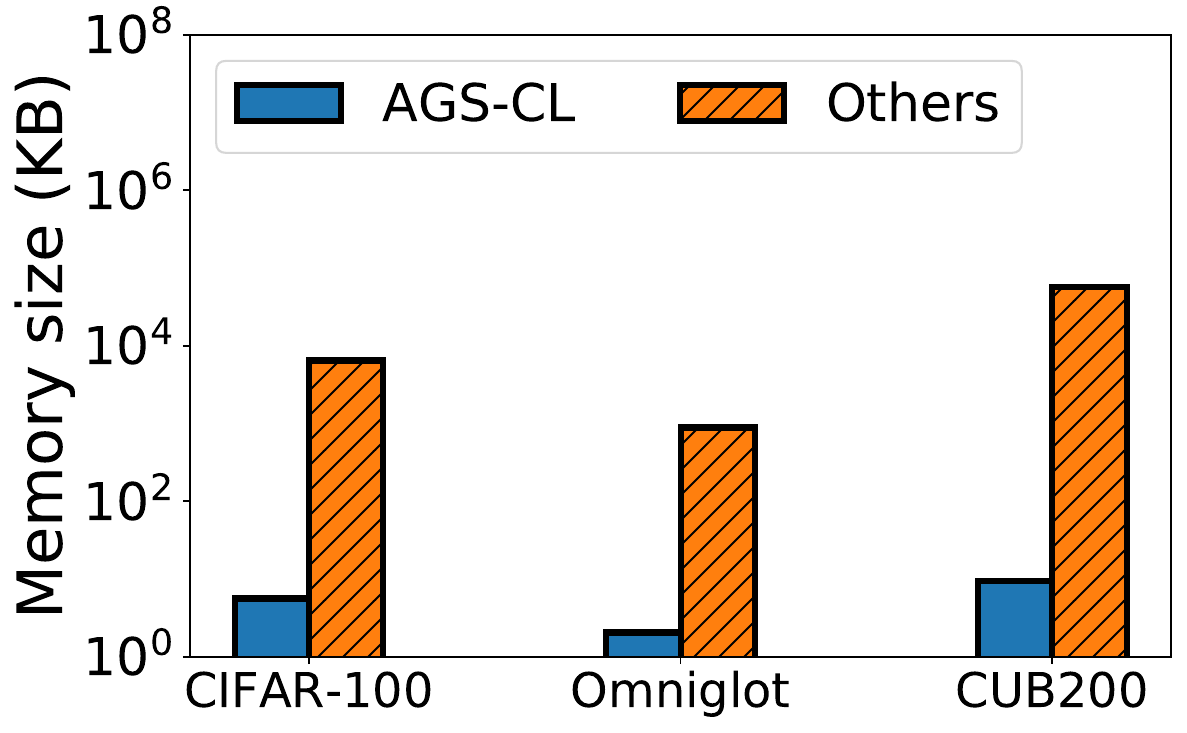}}
    \subfigure[Sparsity and used capacity]{\label{fig:Sparsity_Freeze}
    \includegraphics[width=0.32\columnwidth]{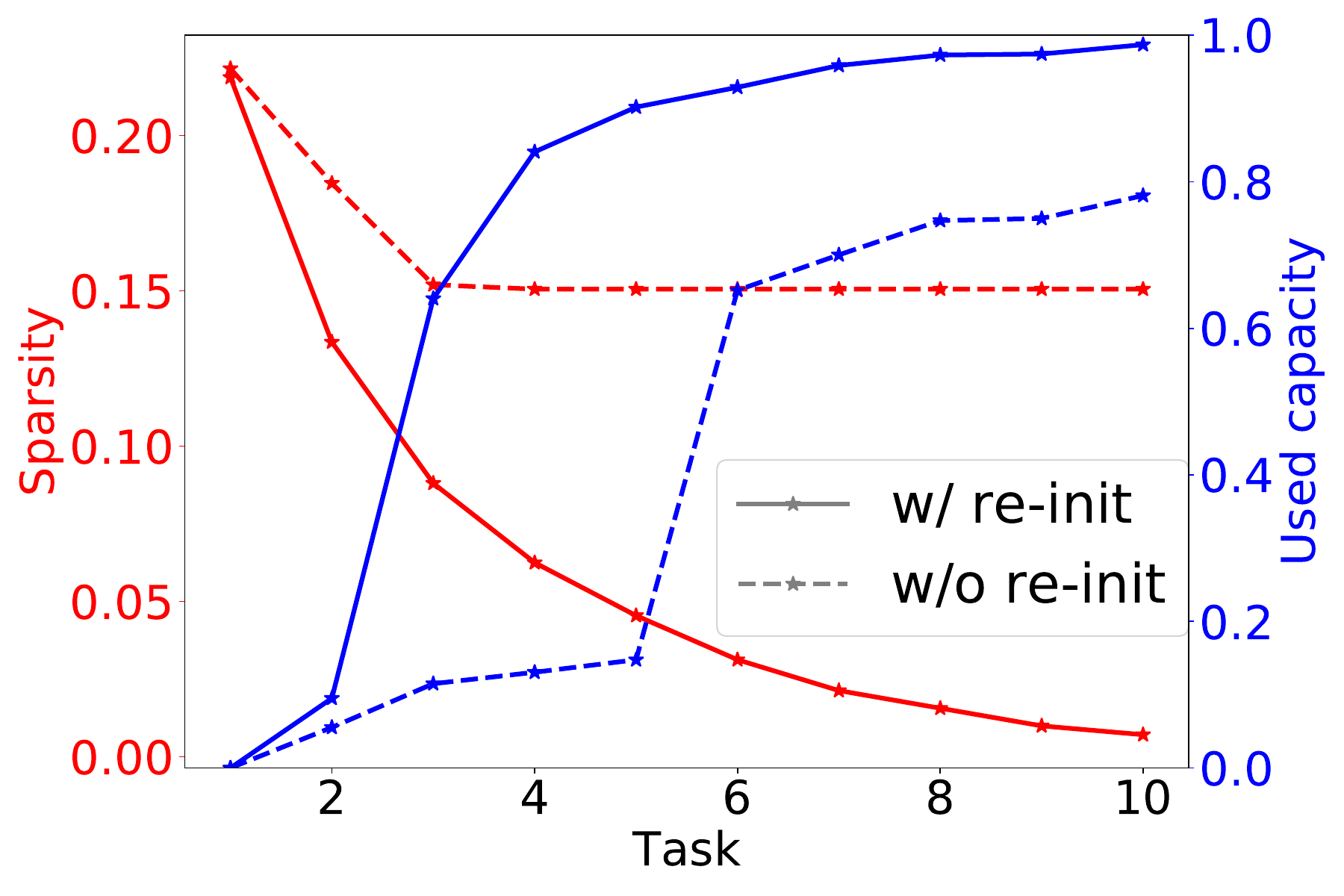}}
    \subfigure[Plasticity ($\mathcal{P}$) and stability ($\mathcal{S}$)]{\label{fig:init_ablation}
    \includegraphics[width=0.32\columnwidth]{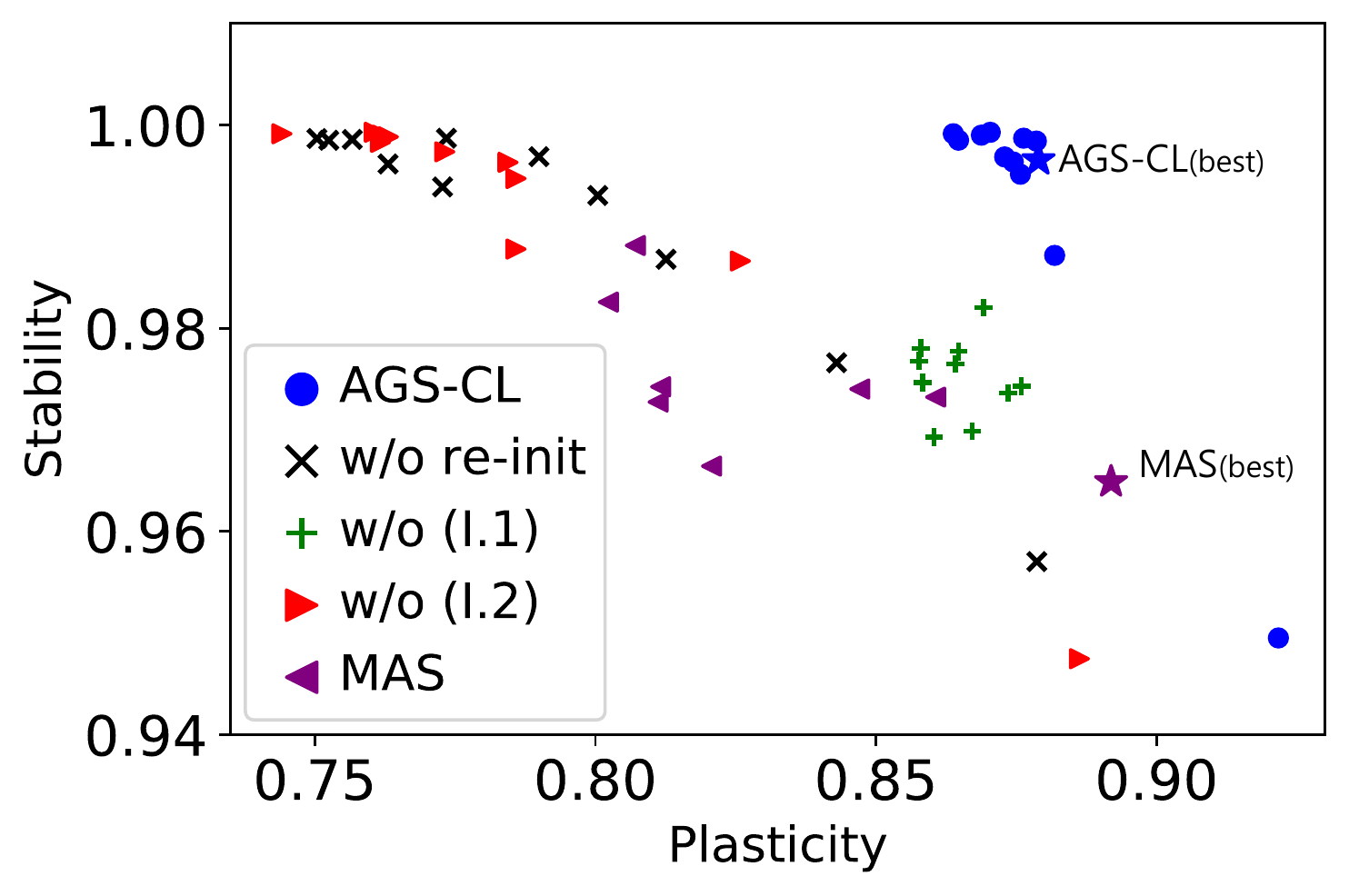}}
    \vspace{-.07in}
    \caption{Various analyses on AGS-CL. (b) and (c) are for CIFAR-100. We note the decreasing and increasing curves in (b) represent sparsity and used capacity, respectively.
    }\label{fig:analysis}    
\end{figure}

\noindent\textbf{Analysis 1: Required memory size} \ \ Figure \ref{fig:Parameters} compares the required memory sizes to store regularization parameters between AGS-CL and other weight-wise regularization methods (\textit{e.g.}, MAS). Note AGS-CL only needs to store the parameters for the nodes, $\{\Omega_{n_{\ell}}^t\}$, whereas others need to store the parameters for the weights. From the figure, we clearly observe that AGS-CL uses \textit{orders of magnitude} less memory than other methods. Particularly, for CUB200, in which a large-scale model (AlexNet) is used, the gap is more than four orders, and such drastic compactness in additional memory gives a competitive edge on the practicality of our AGS-CL. 
\vspace{.02in}\\
\noindent\textbf{Analysis 2: Sparsity and used capacity of the model}\ \ 
Figure \ref{fig:Sparsity_Freeze} closely analyzes how sparsity, the left (red) $y$-axis, and used capacity, the right (blue) $y$-axis, of the network evolve as the learning with AGS-CL continues, for CIFAR-100. 
Moreover, the solid and dashed lines represent the schemes with or without the re-initialization steps, \textit{i.e.}, (I.1) and (I.2), respectively, and we set $\rho=0.3$ for (I.2).
We define the sparsity and the used capacity of a network as the ratios
$$
\frac{|\Gcalz^{t}|}{|\Gcal|} \ \ \ \text{and} \ \ \ \frac{|\{n_\ell: \|\hat{\btheta}_{n_\ell}^{(t)}-\hat{\btheta}_{n_\ell}^{(t-1)}\|_2=0\}|}{|\Gcal|}\ ,
$$
respectively.
Thus, large sparsity implies many ``active learners'' are available for learning future tasks, and large used capacity means many nodes are frozen to not forget past tasks. 
We observe the sparsity and the used capacity gradually decreases and increases, respectively, automatically controlled by $\{\Omega_{n_\ell}^t\}$ and PGD as intended.
We further observe that the re-initialization steps are essential for AGS-CL; without the re-initialization, the network sparsity does not drop beyond a certain level, hence, AGS-CL cannot utilize the full capacity of the network.\\

\noindent\textbf{Analysis 3: Effect of re-initializations}\ \ 
To further study the effectiveness of re-initialization more concretely, we evaluated two additional metrics, plasticity ($\mathcal{P}$) and stability ($\mathcal{S}$).  
To define the metrics, we first let $A \in \mathbb{R}^{\mathcal{T} \times \mathcal{T}}$ be the accuracy matrix of a continual learning algorithm, in which $A_{ij}$ is the accuracy of the $j$-th task after learning the $i$-th task, and let $A^*_i$ be the accuracy of a vanilla fine-tuning scheme for task $i$. Then, the metrics are defined as 
$$
\mathcal{P} \triangleq \frac{1}{\mathcal{T}}\sum_{i=1}^\mathcal{T} \frac{A_{ii}}{A_{i}^*} \ \ \ \text{and} \ \ \ \mathcal{S} \triangleq \frac{1}{\mathcal{T}}\sum_{j=1}^\mathcal{T} \frac{A_{\mathcal{T}j}}{\max_{j \leq i \leq \mathcal{T}} (A_{ij})}\ ,$$
in which $\mathcal{P}$ measures the amount of ``forward transfer'' and $\mathcal{S}$ measures the amount of ``not forgetting'' (\textit{i.e.}, higher the better for both).  
Figure \ref{fig:init_ablation} reports the trade-offs between $\mathcal{P}$ and $\mathcal{S}$, obtained from CIFAR-100 for several variants of AGS-CL and a representative baseline, MAS. For AGS-CL, we ablated each re-initialization scheme in Section \ref{subsubsec:re_init}; `\textit{w/o} (I.1)' is without (I.1) step, `\textit{w/o} (I.2)' is without (I.2) step, and `\textit{w/o} re-init' is without both. Moreover, the plotted trade-offs are over the hyperparameters; \ie, for AGS-CL, we fixed $(\mu,\rho)=(10,0.3)$ and varied $\lambda$, and for MAS, we varied $\lambda$. 
The two `$\star$' points in the figure represent the results of the optimum $\lambda$ for AGS-CL (blue) and MAS (purple). Followings are our observations. First, we clearly see AGS-CL has much better $\mathcal{P}$-$\mathcal{S}$ trade-off than MAS. Namely, AGS-CL hardly suffers from any forgetting (\textit{i.e.}, $\mathcal{S}\approx 1$) and has higher $\mathcal{P}$ values than MAS for most cases. Second, we clearly observe (I.1) improves stability, by comparing AGS-CL and `\textit{w/o} (I.1)' at similar $\mathcal{P}$, and (I.2) improves plasticity, by comparing AGS-CL and `\textit{w/o} (I.2)' at similar $\mathcal{S}$. Finally,  `\textit{w/o} re-init' and `\textit{w/o} (I.2)' show similar performance, hence, (I.1) alone is not enough for attaining both high $\mathcal{P}$ and $\mathcal{S}$.

\subsection{Reinforcement learning on Atari tasks}

We now evaluate the performance of AGS-CL on Atari \cite{(gym)brockman2016openai} reinforcement learning (RL) tasks. As mentioned in the Introduction, a few previous works \cite{(EWC)KirkPascRabi17,(ProgressCompress)SchwarzLuketinaHadsell18} also considered the continual learning of Atari tasks, but their settings allowed the agent to learn past tasks again in a \textit{recurring} fashion. In contrast, we consider \textit{pure} continual learning setting, namely, the past tasks cannot be learned again, but the average rewards are evaluated for all tasks learned so far after learning each task. We randomly selected eight Atari tasks, \textit{i.e.}, \textit{$\{$StarGunner - Boxing - VideoPinball - Crazyclimber - Gopher - Robotank - DemonAttack - NameThisGame$\}$}, and compared AGS-CL with three baselines, EWC, MAS and fine-tuning. 
The CNN agent had three convolution layers, one fully connected layer, and 8 separate output layer for each task, and we used PPO \cite{(PPO)SchulmanWlskiKlimov} identically for learning the agent for all comparing methods. 
Each task is learned with $10^7$ steps, and we evaluated the reward of the agent 30 times per task. We did fair hyperparameter search for all
methods and report the best result for each method.  More detailed experimental settings and are given in Supplementary Materials.  

\begin{figure}[ht]
    \centering
    \subfigure[Normalized accumulated reward]{\label{fig:rl_cumulative}
    \includegraphics[width=0.32\textwidth]{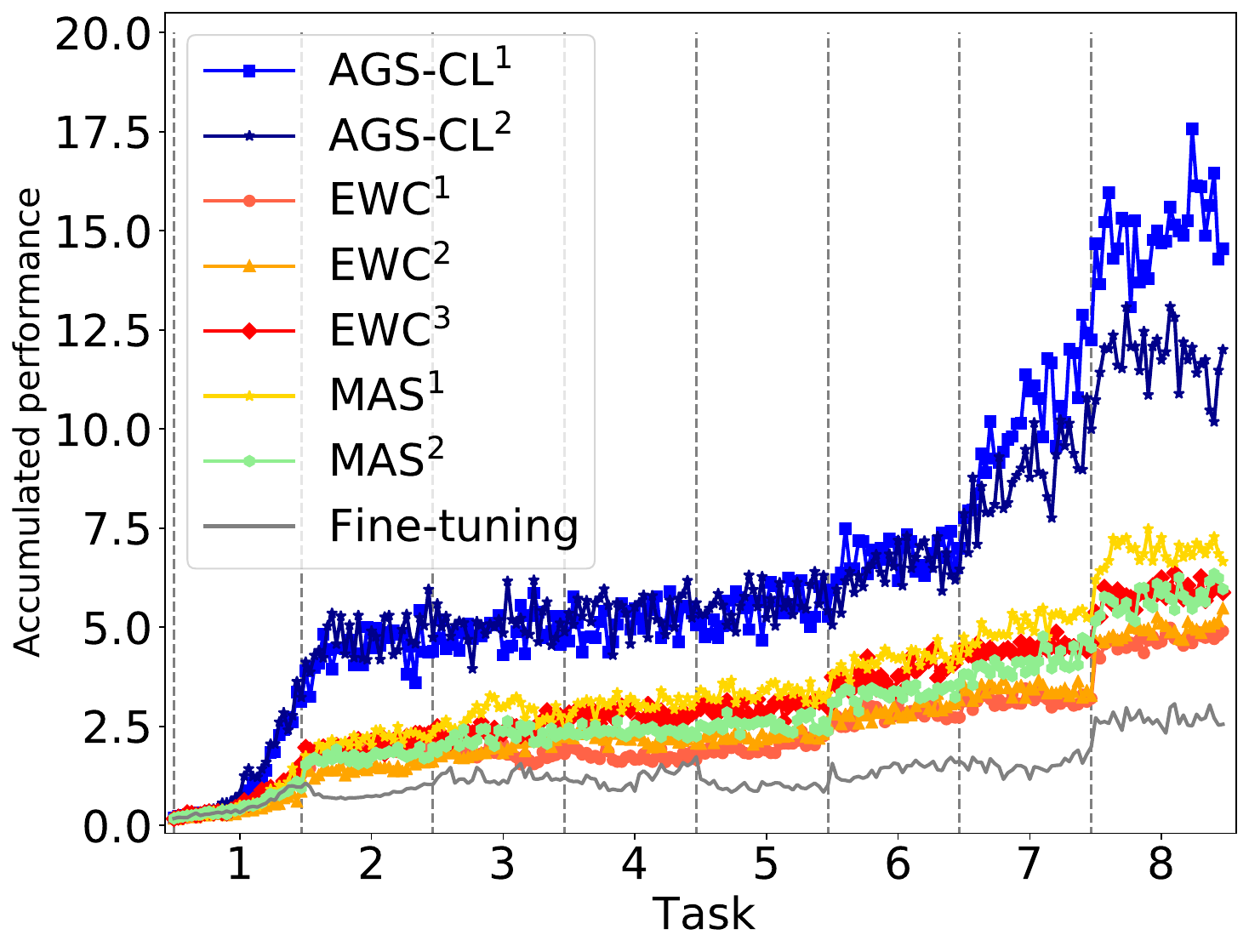}}
    \subfigure[Plasticity ($\mathcal{P}$) and stability ($\mathcal{S}$)]{\label{fig:rl_s_p}
    \includegraphics[width=0.32\columnwidth]{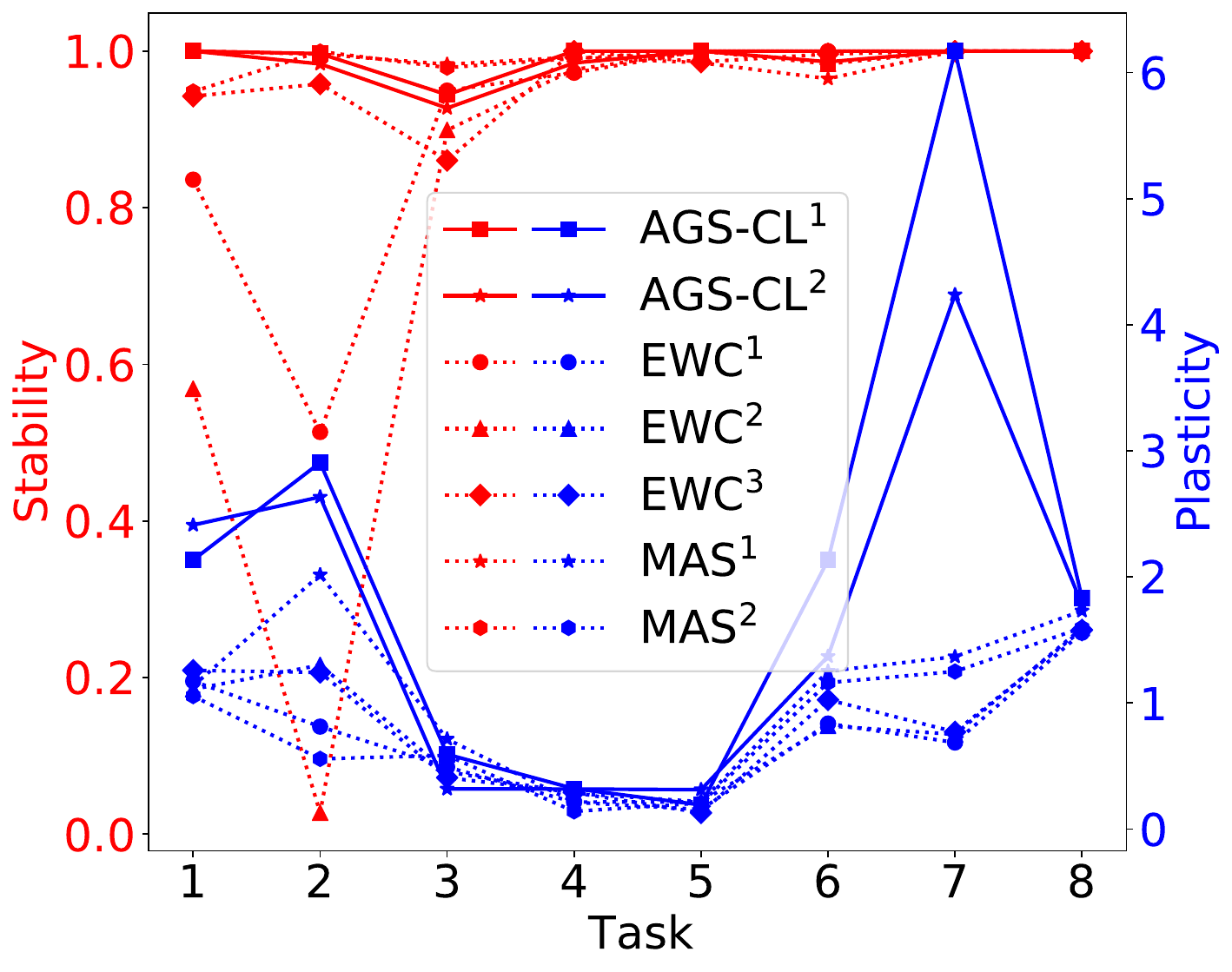}}
    \caption{Reinforcement learning results. $\lambda=\{1,2.5,10\}\times10^4$ for EWC$^{1,2,3}$, $\lambda=\{1,10\}$ for MAS$^{1,2}$, and $\mu= 0.1$, $\lambda=\{1,10\}\times 10^2$ for AGS-CL$^{1,2}$ were used, respectively.  }\label{fig:analysis}
\end{figure}
%


Figure \ref{fig:rl_cumulative} shows the normalized accumulated rewards, in which each evaluated reward is normalized with the maximum reward obtained by fine-tuning for each task, for 8 tasks obtained by the baselines and AGS-CL. 
We clearly observe that AGS-CL achieves much superior accumulated reward at the end of the 8 tasks compared to both EWC and MAS ($\sim3\times$) and fine-tuning ($\sim5\times$).  
Figure \ref{fig:rl_s_p} further considers the plasticity and stability for each task (instead of the average values as in supervised learning). We note AGS-CL hardly suffers from catastrophic forgetting (\textit{i.e.}, stability $\approx 1$ for most tasks) and also does a much better job in learning new tasks than not only the EWC and MAS, but also the fine-tuning (\textit{i.e}, plasticity $\gg1$, particularly for tasks 1,2 and 7).

%% file: conclusion.tex
\section{Concluding Remark}
We proposed AGS-CL, a new continual learning method based on node-wise importance regularization. With a novel loss function based on group-sparsity norms, PGD optimization technique, and the re-initialization tricks, we showed our AGS-CL dominated other state-of-the-arts on various benchmark datasets even with \textit{orders of magnitude} smaller number of regularization parameters. We also show the promixing results on \textit{pure} continual reinforcement learning with Atari tasks.Our method can be also naturally extended to dynamic architecture-based method by simply adding some more free nodes to $\Gcal_0^t$ when the network capacity depletes, which we will pursue as a future work. 

\section{Broader Impact}

We tackle a fairly general continual learning problem, and there is no particular application forseen. The potential societal impact of our work, however, lies in saving intensive usage of computing resources, which is known to affect climate change and global warming due to the excessive energy consumption and necessity of cooling systems. Namely, when numerous ML applications require repetitive re-training of computationally intensive neural networks for learning every new task, overloading of data centers is indispensable. Hence, an effective continual learning algorithm, as proposed in our paper, can save such heavy energy consumption without losing the model accuracy. Furthermore, the effective memory usage can be additional benefit for using our method in memory-limited environments, \textit{e.g.}, mobile devices.




\newpage

\section*{Acknowledgements}

This work is supported in part by Institute of Information \& communications Technology Planning  Evaluation (IITP) grant funded by the Korea government (MSIT) [No.2016-0-00563, Research on adaptive machine learning technology development for intelligent autonomous digital companion], [No.2019-0-00421, AI Graduate School Support Program (Sungkyunkwan University)], [No.2019-0-01396, Development of framework for analyzing, detecting, mitigating of bias in AI model and training data], and [IITP-2019-2018-0-01798, ITRC Support Program].


%% file: group_sparse_neurips.bbl
\begin{thebibliography}{10}

\bibitem{(UCL)ahn2019uncertainty}
Hongjoon Ahn, Sungmin Cha, Donggyu Lee, and Taesup Moon.
\newblock Uncertainty-based continual learning with adaptive regularization.
\newblock In {\em Advances in Neural Information Processing Systems (NeurIPS)},
  pages 4394--4404, 2019.

\bibitem{(MAS)aljundi2018memory}
Rahaf Aljundi, Francesca Babiloni, Mohamed Elhoseiny, Marcus Rohrbach, and
  Tinne Tuytelaars.
\newblock Memory aware synapses: Learning what (not) to forget.
\newblock In {\em Proceedings of the European Conference on Computer Vision
  (ECCV)}, pages 139--154, 2018.

\bibitem{aljundi2018selfless}
Rahaf Aljundi, Marcus Rohrbach, and Tinne Tuytelaars.
\newblock Selfless sequential learning.
\newblock In {\em International Conference on Learning Representations (ICLR)},
  2018.

\bibitem{alvarez2016learning}
Jose~M Alvarez and Mathieu Salzmann.
\newblock Learning the number of neurons in deep networks.
\newblock In {\em Advances in Neural Information Processing Systems (NeurIPS)},
  pages 2270--2278, 2016.

\bibitem{lrp}
Sebastian Bach, Alexander Binder, Gr{\'e}goire Montavon, Frederick Klauschen,
  Klaus-Robert M{\"u}ller, and Wojciech Samek.
\newblock On pixel-wise explanations for non-linear classifier decisions by
  layer-wise relevance propagation.
\newblock {\em PloS One}, 10(7), 2015.

\bibitem{(gym)brockman2016openai}
Greg Brockman, Vicki Cheung, Ludwig Pettersson, Jonas Schneider, John Schulman,
  Jie Tang, and Wojciech Zaremba.
\newblock Openai gym.
\newblock {\em arXiv preprint arXiv:1606.01540}, 2016.

\bibitem{(spdilemma)carpenter87}
Gail~A Carpenter and Stephen Grossberg.
\newblock Art 2: Self-organization of stable category recognition codes for
  analog input patterns.
\newblock {\em Applied Optics}, 26(23):4919--4930, 1987.

\bibitem{(Rwalk)chaudhry2018riemannian}
Arslan Chaudhry, Puneet~K Dokania, Thalaiyasingam Ajanthan, and Philip~HS Torr.
\newblock Riemannian walk for incremental learning: Understanding forgetting
  and intransigence.
\newblock In {\em Proceedings of the European Conference on Computer Vision
  (ECCV)}, pages 532--547, 2018.

\bibitem{(Imagenet)Deng09}
Jia Deng, Wei Dong, Richard Socher, Li-Jia Li, Kai Li, and Li~Fei-Fei.
\newblock Imagenet: A large-scale hierarchical image database.
\newblock In {\em In Proceedings of the IEEE Conference on Computer Vision and
  Pattern Recognition (CVPR)}, pages 248--255, 2009.

\bibitem{(Npruning)golkar2019continual}
Siavash Golkar, Michael Kagan, and Kyunghyun Cho.
\newblock Continual learning via neural pruning.
\newblock {\em Advances in Neural Information Processing Systems (NeurIPS)
  Workshop}, 2019.

\bibitem{(Fearnet)kemker17}
Ronald Kemker and Christopher Kanan.
\newblock Fearnet: Brain-inspired model for incremental learning.
\newblock In {\em International Conference on Learning Representations (ICLR)},
  2018.

\bibitem{KinBa15}
Diederick~P Kingma and Jimmy Ba.
\newblock Adam: A method for stochastic optimization.
\newblock In {\em International Conference on Learning Representations (ICLR)},
  2015.

\bibitem{(EWC)KirkPascRabi17}
James Kirkpatrick, Razvan Pascanu, Neil Rabinowitz, Joel Veness, Guillaume
  Desjardins, Andrei~A. Rusu, Kieran Milan, John Quan, Tiago Ramalho, Agnieszka
  Grabska-Barwinska, Demis Hassabis, Claudia Clopath, Dharshan Kumaran, and
  Raia Hadsell.
\newblock Overcoming catastrophic forgetting in neural networks.
\newblock {\em Proceedings of the National Academy of Sciences},
  114(13):3521--3526, 2017.

\bibitem{(cifar)krizhevsky2009learning}
Alex Krizhevsky, Geoffrey Hinton, et~al.
\newblock Learning multiple layers of features from tiny images.
\newblock 2009.

\bibitem{(alexnet)krizhevsky2012imagenet}
Alex Krizhevsky, Ilya Sutskever, and Geoffrey~E Hinton.
\newblock Imagenet classification with deep convolutional neural networks.
\newblock In {\em Advances in Neural Information Processing Systems (NeurIPS)},
  pages 1097--1105, 2012.

\bibitem{(Omniglot)Lake11}
Brenden Lake, Ruslan Salakhutdinov, Jason Gross, and Joshua Tenenbaum.
\newblock One shot learning of simple visual concepts.
\newblock In {\em Proceedings of the Annual Meeting of the Cognitive Science
  Society}, volume~33, 2011.

\bibitem{(PruningEfficient)Li2016}
Hao Li, Asim Kadav, Igor Durdanovic, Hanan Samet, and Hans~Peter Graf.
\newblock Pruning filters for efficient convnets.
\newblock In {\em International Conference on Learning Representations (ICLR)},
  2017.

\bibitem{(LwF)LiHoiem16}
Zhizhong Li and Derek Hoiem.
\newblock Learning without forgetting.
\newblock {\em IEEE Transactions on Pattern Analysis and Machine Intelligence},
  40(12):2935--2947, 2017.

\bibitem{(LearningEfficientConvNet)Liu2017}
Zhuang Liu, Jianguo Li, Zhiqiang Shen, Gao Huang, Shoumeng Yan, and Changshui
  Zhang.
\newblock Learning efficient convolutional networks through network slimming.
\newblock In {\em Proceedings of the IEEE International Conference on Computer
  Vision (ICCV)}, pages 2736--2744, 2017.

\bibitem{(GEM)LopezRanzato17}
David Lopez-Paz and Marc~Aurelio Ranzato.
\newblock Gradient episodic memory for continual learning.
\newblock In {\em Advances in Neural Information Processing System (NIPS)},
  pages 6467--6476. 2017.

\bibitem{(spdilemma)mermillod13}
Martial Mermillod, Aur{\'e}lia Bugaiska, and Patrick Bonin.
\newblock The stability-plasticity dilemma: Investigating the continuum from
  catastrophic forgetting to age-limited learning effects.
\newblock {\em Frontiers in Psychology}, 4:504, 2013.

\bibitem{(VCL)NguLiBuiTurner18}
Cuong~V. Nguyen, Yingzhen Li, Thang~D. Bui, and Richard~E. Turner.
\newblock Variational continual learning.
\newblock In {\em International Conference on Learning Representations (ICLR)},
  2018.

\bibitem{parikh2014proximal}
Neal Parikh, Stephen Boyd, et~al.
\newblock Proximal algorithms.
\newblock {\em Foundations and Trends{\textregistered} in Optimization},
  1(3):127--239, 2014.

\bibitem{(Review)PariKemPartKananWerm18}
German~Ignacio Parisi, Ronald Kemker, Jose~L. Part, Christopher Kanan, and
  Stefan Wermter.
\newblock Continual lifelong learning with neural networks: {A} review.
\newblock {\em CoRR}, abs/1802.07569, 2018.

\bibitem{(Pytorch)paszke2017automatic}
Adam Paszke, Sam Gross, Soumith Chintala, Gregory Chanan, Edward Yang, Zachary
  DeVito, Zeming Lin, Alban Desmaison, Luca Antiga, and Adam Lerer.
\newblock Automatic differentiation in pytorch.
\newblock {\em Advances in Neural Information Processing Systems (NeurIPS)
  Workshop}, 2017.

\bibitem{(icarl)rebuffi17}
Sylvestre-Alvise Rebuffi, Alexander Kolesnikov, Georg Sperl, and Christoph~H
  Lampert.
\newblock icarl: Incremental classifier and representation learning.
\newblock In {\em Proceedings of the IEEE Conference on Computer Vision and
  Pattern Recognition (CVPR)}, pages 2001--2010, 2017.

\bibitem{(PNN)RusuRabiDesjSoyeKirk2016}
Andrei~A Rusu, Neil~C Rabinowitz, Guillaume Desjardins, Hubert Soyer, James
  Kirkpatrick, Koray Kavukcuoglu, Razvan Pascanu, and Raia Hadsell.
\newblock Progressive neural networks.
\newblock {\em arXiv preprint arXiv:1606.04671}, 2016.

\bibitem{samek2016evaluating}
Wojciech Samek, Alexander Binder, Gr{\'e}goire Montavon, Sebastian Lapuschkin,
  and Klaus-Robert M{\"u}ller.
\newblock Evaluating the visualization of what a deep neural network has
  learned.
\newblock {\em IEEE Transactions on Neural Networks and Learning Systems},
  28(11):2660--2673, 2016.

\bibitem{scardapane2017group}
Simone Scardapane, Danilo Comminiello, Amir Hussain, and Aurelio Uncini.
\newblock Group sparse regularization for deep neural networks.
\newblock {\em Neurocomputing}, 241:81--89, 2017.

\bibitem{(PPO)SchulmanWlskiKlimov}
John Schulman, Filip Wolski, Prafulla Dhariwal, Alec Radford, and Oleg Klimov.
\newblock Proximal policy optimization algorithms.
\newblock {\em arXiv preprint arXiv:1707.06347}, 2017.

\bibitem{(ProgressCompress)SchwarzLuketinaHadsell18}
Jonathan Schwarz, Wojciech Czarnecki, Jelena Luketina, Agnieszka
  Grabska-Barwinska, Yee~Whye Teh, Razvan Pascanu, and Raia Hadsell.
\newblock Progress \& compress: A scalable framework for continual learning.
\newblock In {\em International Conference on Machine Learning (ICML)}, pages
  4528--4537, 2018.

\bibitem{grad-cam17}
Ramprasaath~R Selvaraju, Michael Cogswell, Abhishek Das, Ramakrishna Vedantam,
  Devi Parikh, and Dhruv Batra.
\newblock Grad-cam: Visual explanations from deep networks via gradient-based
  localization.
\newblock In {\em Proceedings of the IEEE International Conference on Computer
  Vision (ICCV)}, pages 618--626, 2017.

\bibitem{SerraSurisMironKarat2018(HAT)}
Joan Serra, Didac Suris, Marius Miron, and Alexandros Karatzoglou.
\newblock Overcoming catastrophic forgetting with hard attention to the task.
\newblock In {\em International Conference on Machine Learning (ICML)}, pages
  4548--4557, 2018.

\bibitem{(DGR)ShinLeeKimKim17}
Hanul Shin, Jung~Kwon Lee, Jaehong Kim, and Jiwon Kim.
\newblock Continual learning with deep generative replay.
\newblock In {\em Advances in Neural Information Processing System (NeurIPS)},
  pages 2990--2999. 2017.

\bibitem{(GroupLasso)wang2008note}
Hansheng Wang and Chenlei Leng.
\newblock A note on adaptive group lasso.
\newblock {\em Computational statistics \& data analysis}, 52(12):5277--5286,
  2008.

\bibitem{(cub)welinder2010caltech}
Peter Welinder, Steve Branson, Takeshi Mita, Catherine Wah, Florian Schroff,
  Serge Belongie, and Pietro Perona.
\newblock Caltech-ucsd birds 200.
\newblock 2010.

\bibitem{wen2016learning}
Wei Wen, Chunpeng Wu, Yandan Wang, Yiran Chen, and Hai Li.
\newblock Learning structured sparsity in deep neural networks.
\newblock In {\em Advances in Neural Information Processing Systems (NeurIPS)},
  pages 2074--2082, 2016.

\bibitem{yoon2017combined}
Jaehong Yoon and Sung~Ju Hwang.
\newblock Combined group and exclusive sparsity for deep neural networks.
\newblock In {\em International Conference on Machine Learning (ICML)}, pages
  3958--3966, 2017.

\bibitem{(DEN)YoonYangLeeHwang18}
Jaehong Yoon, Eunho Yang, Jeongtae Lee, and Sung~Ju Hwang.
\newblock Lifelong learning with dynamically expandable networks.
\newblock In {\em International Conference on Learning Representations (ICLR)},
  2018.

\bibitem{yuan2006model}
Ming Yuan and Yi~Lin.
\newblock Model selection and estimation in regression with grouped variables.
\newblock {\em Journal of the Royal Statistical Society: Series B (Statistical
  Methodology)}, 68(1):49--67, 2006.

\bibitem{(SI)ZenkePooleGang17}
Friedemann Zenke, Ben Poole, and Surya Ganguli.
\newblock Continual learning through synaptic intelligence.
\newblock In {\em International Conference on Machine Learning (ICML)}, pages
  3987--3995, 2017.

\bibitem{(DiscriminationPruning)Zhuang2018}
Zhuangwei Zhuang, Mingkui Tan, Bohan Zhuang, Jing Liu, Yong Guo, Qingyao Wu,
  Junzhou Huang, and Jinhui Zhu.
\newblock Discrimination-aware channel pruning for deep neural networks.
\newblock In {\em Advances in Neural Information Processing Systems (NeurIPS)},
  pages 875--886, 2018.

\bibitem{Zou06}
H.~Zou.
\newblock The adaptive {L}asso and its oracle properties.
\newblock {\em Journal of the American Statistical Association},
  101:1418--1429, 2006.

\end{thebibliography}


\begin{thebibliography}{1}

\bibitem{KinBa15}
Diederick~P Kingma and Jimmy Ba.
\newblock Adam: A method for stochastic optimization.
\newblock In {\em International Conference on Learning Representations (ICLR)},
  2015.

\bibitem{(atari)mnih2013playing}
Volodymyr Mnih, Koray Kavukcuoglu, David Silver, Alex Graves, Ioannis
  Antonoglou, Daan Wierstra, and Martin Riedmiller.
\newblock Playing atari with deep reinforcement learning.
\newblock {\em arXiv preprint arXiv:1312.5602}, 2013.

\bibitem{(PPO)SchulmanWlskiKlimov}
John Schulman, Filip Wolski, Prafulla Dhariwal, Alec Radford, and Oleg Klimov.
\newblock Proximal policy optimization algorithms.
\newblock {\em arXiv preprint arXiv:1707.06347}, 2017.

\bibitem{SerraSurisMironKarat2018(HAT)}
Joan Serra, Didac Suris, Marius Miron, and Alexandros Karatzoglou.
\newblock Overcoming catastrophic forgetting with hard attention to the task.
\newblock In {\em International Conference on Machine Learning (ICML)}, pages
  4548--4557, 2018.

\end{thebibliography}
